\documentclass[10pt]{article}
\usepackage{amsmath}
\usepackage{amssymb}
\usepackage{subcaption}
\usepackage[labelformat=parens,labelsep=quad,skip=3pt]{caption}
\usepackage{graphicx}
\usepackage{listings}
\usepackage{geometry}
\usepackage[usenames,dvipsnames,svgnames,table]{xcolor}
\usepackage{changes}
\usepackage{lipsum}
\usepackage{a4wide}
\usepackage{listings}
\usepackage{algorithm}
\usepackage{algorithmic}
\usepackage{booktabs}
\usepackage{float}
\floatstyle{plaintop}
\restylefloat{table}

\begin{document}
\title{Uncertainty Quantification of Darcy Flow through Porous Media using Deep Gaussian Process
}

\author{Alireza Daneshkhah$^{1}${\footnote {\small Correspondence: Alireza Daneshkhah, Email: ali.daneshkhah@coventry.ac.uk}}, O. Chatrabgoun$^{2}$\\
	  M. Esmaeilbeigi$^{3}$, T. Sedighi$^{4}$, S. Abolfathi$^{5}$ \\
	{\it {\small $^1$Centre for Data Science, Coventry University, UK}}\\
	{\it {\small $^2$Department of Statistics, Malayer University, Iran}}\\
	{\it {\small $^3$Department of Mathematics, Malayer University, Iran}}\\
	{\it {\small $^4$ Centre for Environment and Agricultural Informatics, Cranfield University, UK}}\\
	{\it {\small $^5$ Warwick Centre for Predictive Modelling, The University of Warwick, UK}}
}
\date{}



 \maketitle

\begin{abstract}
In this paper, we present a computational method on the basis of the non-linear Gaussian process (GP) method, known as deep Gaussian processes (deep GPs) for uncertainty quantification (UQ) and propagation (UP) in modelling of flow through heterogeneous porous media. The method is also used for reducing the dimensionality of the model output space and consequently emulating the highly complex relationship between hydrogeological properties and the reduced order fluid velocity field in a tractable manner. Deep GPs are multi-layer hierarchical generalisations of GPs with multiple, infinitely wide hidden layers. Deep GPs are very efficient models for deep learning and modelling of high-dimensional complex systems by tackling the complexity through several hidden layers connected with non-linear mappings. According to this approach, the hydrogeological data is modelled as the output of a multivariate GP whose inputs are governed by another GP such that each single layer is either a standard GP or the Gaussian process latent variable model (GP-LVM). In this work, a variational approximation framework is used so that the posterior distribution of the model outputs associated to given inputs can be analytically approximated. In contrast to the other dimensionality reduction methods which do not provide any information about the dimensionality of each hidden layer, the proposed method automatically selects the dimensionality of each hidden layer (to prevent the final model becomes excessively complex) and it can be used to propagate uncertainty obtained in each layer across the hierarchy. Using this, dimensionality of the full input space consists of both geometrical parameters of modelling domain and stochastic hydrogeological parameters can be simultaneously reduced without the need for any simplifications generally being assumed for stochastic modelling of subsurface flow problems. It allows estimation of the flow statistics with greatly reduced computational efforts compared to other stochastic approaches such as Monte Carlo (MC).
\end{abstract}

\textbf{keywords}\\
Gaussian process; Uncertainty quantification; Surrogate models; Bayesian Gaussian process latent variable model; Stochastic PDEs; Deep Gaussian Process.
\\
\par
\section{Introduction}\label{sec:Intro}
In realistic subsurface flow problems, inherent heterogeneity of natural porous media, and lack of information about the micro-structure of porous media and their physical properties cause uncertainties in predicting flow and transport processes. To realistically simulate these processes, effects of the heterogeneities should be incorporated into the classical governing equations describing the flow and transport problems, and possible uncertainties should be quantified. For this, the classical partial differential equations need to be solved with some stochastic approaches. Monte-Carlo approach is a powerful method for uncertainties quantification in a system, has been widely used for modelling flow and transport in porous media~\cite{journals/mousavi2011,journalsVole,journals/Guadagnini2004}. In this method, according to specified probabilistic distributions for random properties of porous media, different samples for related parameters of the problem governing equations are generated, and the equations are solved repeatedly for each sample. Therefore, depending on the number of contributing parameters in the problem, the size of the domain under consideration, and magnitude of the sampling errors, Monte-Carlo approach can be computationally intensive and time consuming ~\cite{journals/mousavi2010,farsi2017mathematical}. Analytical-based stochastic approaches which are mainly developed on the basis of perturbation and moment equation methods~\cite{journals/mousavi2011,journals/zhang2004,book/Gelhar}   are other types of modelling techniques that is capable to consider the probabilistic nature of hydrological processes in the porous media. Previous investigations of authors  ~\cite{journals/mousavi2011,journals/mousavi2013,journals/mousavi2011(2)} have shown that the perturbation theory is capable to develop transparent close form stochastic equations describing correlations between stochastic parameters of the modelling domain and physical transport mechanisms. However, the need for utilizing advanced analytical methods to solve stochastic governing equations, makes the approach unfeasible and and cumbersome to be used for more complex flow conditions such as multiphase flow condition.
\par
The alternative method to mitigate aggregate simulation cost is to build an  accurate, cheap to-evaluate surrogate model that can be interrogated in place of expensive simulations~\cite{Owen.et.al.2017}. The surrogate-based UQ methods have received increasing attention over the past two decades, among them Polynomial chaos (PC) and Gaussian process (GP) emulations. The PC and its other variants including generalized PC (gPC) and multi-element gPC (ME-gPC)~\cite{Xiu2002a,Wan2005} suffer from various drawbacks, including modeling of discontinuities in the response surface and the curse of dimensionality~\cite{journals/jcphy/BilionisZ12,journals/jcphy/BilionisZKL13}.
\par
The GP surrogate was originally developed for curve fitting and regression, but it has recently become popular for a wide-range of supervised and unsupervised applications~\cite{rasmussen:williams:2006} including regression, classification, image processing and data visualization. In this method, the simulator is considered as an unknown function of its input parameters, which is modeled as a realization of multivariate normal distribution. 
The GP model was extended for prediction tasks in~\cite{citeulike:5719017} within a fully Bayesian framework. The methodology has been  revisited and refined in~\cite{RSSB:RSSB294} in terms of mathematical elegance, and to incorporate additional sources of uncertainty (including model discrepancy, parametric uncertainty, observation error) so that the methodology can be used in other disciplines. Emulators have then been applied to deal with various objectives, including uncertainty analysis~\cite{Oakley.O'Hagan.2002}, sensitivity analysis~\cite{Oakley.OHagan.2004} and calibration~\cite{RSSB:RSSB294,journals/siamsc/HigdonKCCR04}.
\par
The scalar GP model was extended in~\cite{Conti.OHagan.2010} by placing a multi-dimensional GP prior over the simulator. They assumed a separable covariance structure
which is clearly a restrictive assumption~\cite{doi:10.1080/00401706.2012.715835}.  
Despite this simplification, the multivariate GP models become computationally exhaustive as the dimensionality of the output space increases~\cite{journals/jcphy/BilionisZ12}.
\par
These computational intractability of the multivariate GP models with high-dimensional outputs could be resolved by considering an efficient dimension reduction (DR) approach before applying the emulator~\cite{maniyar2007dimensionality}. Such approaches 
for both the model input and output include techniques such as principal component analysis (PCA), generative topographic mapping, Gaussian process latent variable model (GP-LVM).
In~\cite{citeulike:6556526} PCA was applied to the output space.  
Various non-linear dimensionality reduction approaches have also been developed~\cite{maniyar2007dimensionality}, such as multidimensional scaling, isometric mapping (Isomap), Kernel PCA (KPCA), GP-LVM, and others. The applications of these techniques to UQ tasks remains limited as 
most of these approaches are not fully probabilistic. 
\par
In this paper, we first present a comprehensive approach for dimensionality reduction of the output space and then proceed by emulating the highly complex relationship between input and the dimensionality reduced output in a very tractable manner. The multi-dimensional emulation model is broken into several stages to enhance predictive performance of the emulator as well as its computational tractability. This hierarchical emulator model is constructed by combining the deep neural network (deep NN) and GP-LVM. This new surrogate model which is resulted from nesting GPs is called \textit{deep GP}~\cite{Damianou.2015} due to the relationship between these models and deep neural network (NN) model~\cite{Damianou.Lawrence.2013,Hensman-Lawrence.2014}. Each layer of the selected NN architecture is modeled as the output of a multivariate GP, whose inputs are governed by another GP. However, the final model resulted from the composition of the GPs is no longer a GP, thus more complex interactions between the input/output data can be effectively learned. The intermediate hidden variables can be thought of as the proper coarse grained variables needed to accurately capture the complex (often multiscale) nature of the input/output variables. 
\par
One of the advantages of our proposed methodology is that the full input space consists of both physical domain and stochastic inputs can be simultaneously reduced, and no simplification as was considered in~\cite{journals/jcphy/BilionisZ12} is assumed in this work. 
The deep GP approach to UQ is straightforward, and efficiently overcomes both the out-of-sample problems (projecting a new high-dimensional sample into its low-dimensional representation) and pre-image problems (projecting back from the low-dimensional space to the full-data space). In addition, the proposed methodology provides an effective probabilistic modelling of high dimensional data which lie on a non-linear manifold~\cite{Lawrence.2005}. In addition, a deep GP as a generative model is able to capture non-stationary and non-smoothness in the stochastic space~\cite{Damianou.2015,Hensman-Lawrence.2014}.
One can analytically approximate the  posterior distribution of the high-dimensional model responses given the high-dimensional inputs using a variational methodology. The resulting distribution can be then used to automatically select and tune the model architecture without the need for cross-validation. The approach is demonstrated with classical UQ tasks including 
computing the error bars for the statistics of interest in a number of stochastic PDE based models.
\par
Section~\ref{sec:GP_Reg} reviews the fundamentals of multivariate GP regression needed in this paper. Section~\ref{sec:Methodology} is dedicated to the general methodology of reducing the dimensionality of the input and output fields, and emulating the relationship between these reduced fields.    In Section~\ref{sec:DeepGP}, we show how this variational methodology can be extended in the hierarchical setting and through the latent layers to train the deep GP model. 
We then discuss how the deep GP framework allows for efficient regularization, as well as automatic structure discovery. In this section, we also present numerical algorithms to compute posterior predictive means and variances of each latent layer and model output. We finish this section by discussing how the trained deep GP model can be used for UQ and computing the quantities of interest including error bars on quantities of interest. Section~\ref{sec:uq} provides a numerical example for Darcy flow through porous media. We conclude in Section~\ref{sec:con}.

\section{Unsupervised Learning Using Gaussian Process Models}\label{sec:GP_Reg}

Here, we discuss how GP emulators can be used for modelling nonlinear regression. Since the physical model of interest to this study is complex. The outputs can be derived as the computer code model outputs~\cite{Oakley.O'Hagan.2002,Conti.OHagan.2010} which return a multi-output response $\mathbf{y}\in \mathbb{Y}\subset \mathbb{R}^{\nu}$, where the output space is denoted by $\mathbb{Y}$ and $\nu > 0$ stands for the dimension of $\mathbb{Y}$. The corresponding input to $\mathbf{y}\in \mathbb{Y}$ is illustrated by $\mathbf{x}\in \mathbb{X}\subset\mathbb{R}^{\kappa}$, where the input space, $\mathbb{X}$ dimension is $\kappa>0$. Due to the purpose of our study, it would be plausible to consider $\mathbf{x}\in \mathbb{X}$ as a second-order random field, illustrated by $\mathbf{x}=\mathcal{Q}(s, \lambda)$, where $s\in \mathbb{S}\subset \mathbb{R}^{2}$, $\lambda\in \Lambda$; $\mathbb{S}$ exhibits a bounded spatial domain and $\Lambda$ is a sample space. The inputs (random fields) are generated by sampling from a distribution $\zeta(\mathbf{x})$ where the dependencies between input points over $\mathbb{S}$ (the spatial domain of interest) can be included using a suitable covariance function. A plausible candidate for $\zeta(\mathbf{x})$ is a zero mean GP distribution with the squared-exponential kernel function~\cite{Lord.etal.2014}. The covariance function can be approximated using the Karhunen-Lo\`eve Expansion (KLE) method and illustrate over a grid with the size of $g_{x}\times g_{x}$. Here, we only use the KLE method as a tool to generate realisations from the considered random field. Therefore, the dimension of the input space $\mathbb{X}\subset\mathbb{R}^{\kappa}$ is thus computed as $\kappa=g_{x}\times g_{x}$, and not the number of terms that for example the KL expansion is truncated at.
\par
The output, $\mathbf{y}$ at the input $\mathbf{x}$, is represented by $\mathbf{y}=\mathbf{f}(\mathbf{x})$, and the output is also a random field evaluated over the $g_{y}\times g_{y}$ grid, and $\mathbf{f}: \mathbb{X}\to \mathbb{Y}$ takes the input data $\mathbf{x}\in \mathbb{X}$ for a Partial Differential Equation (PDE) into the solution $\mathbf{y}\in \mathbb{Y}$ as a forward mapping (see Section~\ref{sec:uq} for further details). The main aim of Uncertainty Propagation (UP) is to determine the propagation of the randomness in the input $\mathbf{x}$ into the randomness in some quantity of interest of the output $\mathbf{y}$ including the mean, variance and probability distribution function (pdf). For complex physical models, evaluating the probability density function $\mathbf{f}(.)$ is computationally expensive. Therefore, the aforementioned quantities of interests (QoIs) cannot be evaluated using the conventional computational approaches such as MC methods which demand too many model runs. In order to evaluate the QoIs, $\mathbf{f}(.)$ must be approximated based on the limited number of the model runs in a computationally and mathematically feasible manner.   
\par
In other words, we wish to learn about $\mathbf{f}(\cdot)$ such that its point estimate and uncertainty (e.g., error bars) around the point estimate can be efficiently derived based on the limited observed data, formulated as $\mathcal{D}=(\mathbb{X}, \mathbb{Y})$. Here $\mathbb{X}=(\mathbf{x}_{1},\ldots,\mathbf{x}_{n})^{T}\in \mathbb{R}^{n\times \kappa}$ {are the observed inputs, and $\mathbb{Y}=(\mathbf{y}_{1},\ldots,\mathbf{y}_{n})^{T}\in \mathbb{R}^{n\times \nu}$ are the corresponding outputs. In this regard, $\mathbf{y}_{i}$ is a random field realisation evaluated over the $g_{y}\times g_{y}$ gird, reshaped in a column vector of size $\kappa=g_{y}\times g_{y}$ and obtained by solving the governing PDE at $\mathbf{x}_{i}$. A plausible way to probabilistically estimate $\mathbf{f}(\cdot)$ given small number of observation is to follow the Bayesian paradigm originally proposed in~\cite{Oakley.O'Hagan.2002}, where $\mathbf{f}(\cdot)$ is considered as an unknown function, and a GP distribution is placed as a prior over functional space. In this context, a stochastic process is called \textit{Gaussian Process} if any finite subset, $\mathcal{F}(\mathbb{X})=(\mathbf{f}(\mathbf{x}_1), \mathbf{f}(\mathbf{x}_2), \ldots \mathbf{f}(\mathbf{x}_n))$ has a joint Gaussian distribution with mean vector $\mbox{\boldmath{$\mu$}}(\mathbb{X})$ and the covariance matrix $\mathcal{K}(\mathbb{X}, \mathbb{X})$. Since uncertainty about $\mbox{\boldmath{$\mu$}}(.)$ can be taken into account by adding an extra term to the kernel, we assume that the mean function is simply zero everywhere, i.e., $\mbox{\boldmath{$\mu$}}(.)=\mathbf{0}$. Each element of covariance matriz $\mathcal{K}(.,.)$ is constructed from the correlation  function between the input data points as follows
	\[
	{k}_{ij}=k(\mathbf{x}_{i},\mathbf{x}_{j}),~\forall i\neq j,~i,j=1,\ldots,n.
	\] 
	It should be noted that this correlation function can characterise the correlations of the outputs at the selected input data points.
	\par
	The correlation function which is widely used in many applications and also considered throughout this paper is the exponential quadratic (also known as RBF) with the following functional form
	\begin{equation}
	k(\mathbf{x}_{i}, \mathbf{x}_{j})=\tau^{2}_{0}\exp\{-(\mathbf{x}_{i}-\mathbf{x}_{j})^{T}\mathbf{B} (\mathbf{x}_{i}-\mathbf{x}_{j})\},\label{Kernel_func:GP}
	\end{equation}
	where $(\tau_{0}, \mathbf{B})$ denote the covariance function hyper-parameters,  $\mathbf{B}$ is a diagonal matrix of positive smoothness parameters (or length-scale parameters), $\mathbf{b}=(b_{1},\ldots,b_{\kappa})$. 
	\par
	The output $\mathbf{y}$ will then can be presented in the following form given this assumption that the data are generated with Gaussian white noise $\mbox{\boldmath{$\epsilon$}}$ around the function $\mathbf{f}(.)$, 
	\begin{equation}
	\mathbf{y}=\mathbf{f}(\mathbf{x})+\mbox{\boldmath{$\epsilon$}},~~\mbox{\boldmath{$\epsilon$}}\sim \mathcal{N}(\mathbf{0},\sigma^{2} \mathbf{I}).\label{noisy_GPM}
	\end{equation}
	The likelihood function is then given by  
	\begin{equation}
	p(\mathbb{Y}\mid\mathcal{F}, \mathbb{X})=\mathcal{N}(\mathcal{F}, \sigma^{2}\mathbf{I}).\label{GP:likelihood1}
	\end{equation}
	where $\mathcal{F}=(\mathbf{f}(\mathbf{x}_1), \mathbf{f}(\mathbf{x}_2), \ldots \mathbf{f}(\mathbf{x}_n))$ follows a GP with the above mean and covariance functions, that is,
	\begin{align}
	p(\mathcal{F}\mid \mathbf{X})=\mathcal{N}(\mathbf{0}, \mathcal{K}(\mathbb{X}, \mathbb{X})),\label{GP_LatentF}
	\end{align}
	\par
	{The joint distribution of $(\mathbb{Y},\mathcal{F})$ given $\mathbb{X}$ can be obtained by multiplying Eqs.~\ref{GP_LatentF} and \ref{GP:likelihood1}. In order to having a modelling framework where a whole family of functions can be simultaneously considered, we need to integrate out the function variables $\mathcal{F}$ which results in
		\begin{equation}
		p(\mathbb{Y}\mid \mathbb{X})=\int{p(\mathbb{Y}\mid \mathcal{F}, \mathbb{X})p(\mathcal{F}\mid \mathbb{X})d\mathcal{F}}=\mathcal{N}(\mathbf{0},\mathcal{K}(\mathbb{X},\mathbb{X})+ \sigma^{2}\mathbf{I}).\label{GP:likelihood2}
		\end{equation}
		This framework can improve the model fitting through the optimisation of $p(\mathbb{Y}\mid \mathbb{X})$ with respect to $(\sigma, \tau_{0}, \mathbf{b})$ which results in the maximum likelihood estimates. 
		\par
		In order to predict the latent variables at arbitrary test input locations $\mathbf{x}^{*}$ based on the noisy observations, we can write the predictive distribution of the GP regression model as
		\begin{eqnarray}
		\mathbf{f}^{*}\mid \mathbb{X}, \mathbb{Y},\mathbf{x}^{*}\sim \mathcal{N}(\mbox{\boldmath{$\mu$}}_{f^{*}}, \mbox{\boldmath{$\Sigma$}}_{f^{*}}),\label{pre_dis_GP} 
		\end{eqnarray}
		where
		\begin{eqnarray}
		\mbox{\boldmath{$\mu$}}_{f^{*}}={\mathcal{K}(\mathbf{x}^{*}, \mathbb{X})[\mathcal{K}(\mathbb{X}, \mathbb{X})+\sigma^{2}\mathbf{I}]^{-1}\mathbb{Y}}, \label{pre_mean_GP} \\
		\mbox{\boldmath{$\Sigma$}}_{f^{*}}={\mathcal{K}(\mathbf{x}^{*}, \mathbf{x}^{*})-\mathcal{K}(\mathbf{x}^{*}, \mathbb{X})[\mathcal{K}(\mathbb{X}, \mathbb{X})+\sigma^{2}\mathbf{I}]^{-1}\mathcal{K}(\mathbb{X},\mathbf{x}^{*})}, \label{pre_var_GP} 
		\end{eqnarray}
		where $\mathbf{x}^{*}$ is the test input vector and $\mathbf{f}^{*}$ is the corresponding test function values. Moreover, $\mathcal{K}(\mathbf{x}^{*}, \mathbb{X})$ and $\mathcal{K}(\mathbf{x}^{*}, \mathbf{x}^{*})$ are the covariance matrices evaluated at all pairs of training and test points \cite{rasmussen:williams:2006,journals/jcphy/BilionisZKL13}. The predictive distribution presented above will be used for prediction the test points on the first latent variable and propagating the predicted output through the latent layers of the trained deep GP as explained in Section~\ref{Sec:Pre_MC_GPLVM}. The steps required to compute the predictive mean and covariance can be found in~\cite{rasmussen:williams:2006}.
	\section{The Deep Gaussian Process Methodology}\label{sec:Methodology}

In this section, an alternative for the multivariate GP model is proposed. The proposed surrogate model is applicable for deep learning and modelling of highly nonlinear and high-dimensional responses. This multivariate GP model is referred to as \textit{deep GP} in this study. In this extent, a set of observations is required to construct the surrogate model which can be expressed as $\mathcal{D}=(\mathbb{X}, \mathbb{Y})$, as explained earlier.
Deep GP is constructed based on the Stochastic Process Composition in probability theory which allows us to break down the emulation into several pieces. This approach would ease the computationally complexity of the multivariate GP models for emulating high-dimensional inputs and outputs. In a deep GP model, a multivariate emulator can be constructed through process composition as thus:
\begin{equation}
\textbf{y}=\textbf{f}_{G}(\textbf{f}_{G-1}(\ldots \textbf{f}_{1}(\textbf{x})))+\mbox{\boldmath{$\epsilon$}},\label{nested_GP}
\end{equation}
Each element in the above composition is assumed to be drawn from a GP model as $\textbf{f}_{i} \sim \mathcal{GP}(\textbf{0}, k(.,.))$ for $i=2,\ldots,G$, where $k(.,.)$ is the covariance function. In such a model, the data is modelled as the output of a multivariate GP such that inputs are governed by another GP. Ultimately, the resulting model out of this composition process represent the complex interactions between data which is no longer a GP.
\par
A deep GP can be considered as a graphical model with three type of nodes: the parent nodes $\mathbb{X}$ as observed inputs data; the intermediate latent/hidden layers $\textbf{h}_{i}\in \mathbb{R}^{n\times q_{i}}$, and the observed outputs data $\mathbb{Y}$. In this regard, $\textbf{h}_{i}$ can be defined as,
$\textbf{h}_{i}=\textbf{f}_{i}(\textbf{h}_{i-1})+\mbox{\boldmath{$\epsilon$}}_{i}$, for $i=2,\ldots,G+1$, where $G$ is the number of latent/hidden layers, $\mbox{\boldmath{$\epsilon$}}_{i}\sim N(\textbf{0}, \sigma_{i}^{2}\textbf{I})$, and $\textbf{f}_{i}(.)$ is zero mean GP with specific covariance function, $k(.,.)$. The three type of nodes for the deep GP graphical model can be placed at the leaf nodes ( see Fig~\ref{DGP-sup}).
\par
\subsection{A Brief Introduction to Learning of the Gaussian Process Latent Variable Model}\label{Sec:GP-LVM}
To construct a deep GP model for emulating high-dimensional input and output fields, the initial step is to train the model. In this regard, it is required to determine nonlinear latent variables. The GP-LVM is presented as a flexible unsupervised GP approach for nonlinear dimensionality reduction. This approach is key in determining nonlinear latent variables. Afterwards, an efficient Bayesian variational method is introduced for training GP-LVMs. Suppose the observed output data matrix $\mathbb{Y} \in \mathbb{R}^{n\times \nu}$ and $\mathbb{H} \in \mathbb{R}^{n\times q} $  where $n$ is the number of observations, $\nu$ is the dimensionality of each data vector and $q \ll \nu$ for the purpose of dimensionality reduction. Moreover, $\textbf{h}_{i,:} \in \mathbb{R}^{\nu}$ and $\textbf{h}_{i,:} \in \mathbb{R}^{q}$ are denoted as the $i^{th}$ row of $\mathbb{Y}$ and $\mathbb{H}$, respectively. Therefore, 
the GP-LVM can be defined as a generative mapping from the latent space to the observation space as \cite{Lawrence.2005}:
\begin{equation}
\mathbf{y}_{i,:}={\mathbf{f}(\mathbf{h}_{i,:})}+\mbox{\boldmath{$\epsilon$}}_{i},~~~~ \mbox{\boldmath{$\epsilon$}}_{i}\sim \mathcal{N}(0, \sigma^{2}\mathbf{I}),~~{i=1,\ldots,n},\label{gen-map1}
\end{equation}
\noindent
Using Eq.~\ref{nested_GP}, the multivariable emulator $\mathbf{f}(\mathbf{h}_{i,:})=(f_{1}(\mathbf{h}_{i,:}),\ldots, f_{d}(\mathbf{h}_{i,:}))$ and the zero-mean GP model is thus:
\begin{align}
{{f}_{g}(\mathbf{h}_{i,:})\sim \mathcal{GP}(\mathbf{0}, k(\mathbf{h}_{i,:},\mathbf{h}'_{i,:})),~~~l=1,\ldots,\nu.} \label{f_element}
\end{align}

In order to create a Bayesian framework including uncertainty quantification, It is required to express the covariance function $k(.,.)$ in the form of the squared exponential automatic relevance determination (ARD) kernel with different length scales and considering Bayesian variation GP-LVM \cite{Damianou:variational15}. The 
covariance function can be determined as:
\begin{equation}
k(\mathbf{h}_{i,:}, \mathbf{h}_{j,:})=\sigma_{h}^{2}\exp\left(-\frac{1}{2}\sum_{k=1}^{q}\omega_{k}(\mathbf{h}_{i,k}-\mathbf{h}_{j,k})^{2}\right),\label{Kernel_ARD}
\end{equation}
\noindent
where the covariance hyper-parameters is $\mbox{\boldmath{$\theta$}}_{f}=(\sigma_{h}, \omega_{1},\ldots, \omega_{q})$ and $\omega$ is a common length-scale parameter for all latent space directions. Considering the Bayesian treatment of the GP-LVM, the log marginal likelihood can be  defined as follows

\begin{equation}
\begin{aligned}
\log p(\mathbb{Y})=&\log\int p(\mathbb{Y},\mathcal{F},\mathcal{H})~d\mathcal{H}d\mathcal{F}, \\
=&\int p(\mathbb{Y}\mid \mathcal{F})p(\mathcal{F}\mid \mathcal{H})p(\mathcal{H})~d\mathcal{H}d\mathcal{F}, \\
=&\log \int p(\mathbb{Y}\mid \mathcal{F})\left(\int p(\mathcal{F}\mid \mathcal{H})p(\mathcal{H})~d\mathcal{H}\right)d\mathcal{F},\label{log_Y_2ndIntg}
\end{aligned}
\end{equation}
\noindent
where, $\mathcal{H}$ is the input data matrix which can be treated as the latent variables and proportionate $p(\mathcal{H})$ through the nonlinear mapping is thus:
$$
\int p(\mathcal{H})\prod_{j=1}^{\nu} p(\mathbf{f}_{;,j}\mid \mathcal{H})~d\mathcal{H},
$$
where $p(\mathbf{f}_{;,j}\mid \mathcal{H})$ is proportional to 
$
|\mathcal{K}_{ff}|^{-\frac{1}{2}}\exp\left(-\frac{1}{2}\mathbf{f}_{:,j}^{T}\mathcal{K}_{ff}^{-1}\mathbf{f}_{:,j}\right)
$
and $\mathbf{f}_{:,j}$ is the $j^{th}$ column of $\mathcal{F}$, and $\mathcal{K}_{ff}=k(\mathcal{H}, \mathcal{H})$ is the covariance matrix derived based on the kernel function given in Eq.~(\ref{Kernel_ARD}). Based on Jensen's inequality the variational lower bound for $\log p(\mathbb{Y})$ can be expressed as:
\begin{equation}
\begin{aligned}
\log p(\mathbb{Y})\geq\log &\int q(\mathcal{F})q(\mathcal{H})\log p(\mathcal{F}\mid \mathcal{H})~d\mathcal{F}d\mathcal{H}~+ \\
&\int q(\mathcal{F})q(\mathcal{H})\frac{p(\mathbb{Y}\mid \mathcal{F})p(\mathcal{H})}{q(\mathcal{F})q(\mathcal{H})}~d\mathcal{F}d\mathcal{H} \label{LoWB_Y}
\end{aligned}
\end{equation}
\noindent
The latent input elements of $\mathcal{H}$ appeared nonlinearly in the inverse kernel matrix, $\mathcal{K}_{ff}^{-1}$ of $\log p(\mathbb{Y})$. Therefore, the analytical integration over $\mathcal{H}$ is infeasible~\cite{Bishop:2006:PRM:1162264}. This limitation can be overcome using sparse approximation methods~\cite{conf/nips/SnelsonG05,journals/jmlr/Titsias09}. Regarding this, an approximation can be constructed based on a small set of $m$ ($\ll n$) auxiliary variables, that allow the reduction of the time complexity from $\mathcal{O}(n^{3})$ to $\mathcal{O}(nm^{2})$.  The main idea behind these methods is to augment the GP model with the pairs of input and output variables, denoted by $\{(\mathbf{h}_{u})_{i,:}\}_{i=1}^{m}$ and $\{\mathbf{u}_{i,:}\}_{i=1}^{m}$, respectively. These variables are stored in matrices, $\mathcal{H}_{u}\in \mathbb{R}^{m\times q}$ and $\mathcal{U}\in \mathbb{R}^{m\times \nu}$, where $\mathcal{H}_u$ is the inducing inputs matrix and $\mathcal{U}$ is the inducing output matrix. The marginal GP prior over the inducing variables can be derived using GP mapping \cite{Damianou.2015} as:
\begin{equation}
\begin{aligned}
p&(\mathcal{U}\mid \mathcal{H}_{u})=\prod_{j=1}^{d}p(\mathbf{u}_{:,j}\mid \mathcal{H}_{u}),~~\textrm{where} \\
p&(\mathbf{u}_{:,j}\mid \mathcal{H}_{u})=\mathcal{N}(\mathbf{u}_{:,j}\mid \mathbf{0},\mathcal{K}_{uu}).\label{prior_marg_indc}
\end{aligned}
\end{equation}
\noindent
Moreover, the conditional GP prior can be presented as:
\begin{equation}
\begin{aligned}
p &(\mathcal{F}\mid \mathcal{U},\mathcal{H}, \mathcal{H}_{u})=\prod_{j=1}^{\nu}p(\mathbf{f}_{:,j}\mid \mathbf{u}_{:,j}, \mathcal{H}, \mathcal{H}_{u}),~~\textrm{where} \\
p &(\mathbf{f}_{:,j}\mid \mathbf{u}_{:,j}, \mathcal{H}, \mathcal{H}_{u})=\mathcal{N}(\mathbf{f}_{:,j}\mid \mbox{\boldmath{$\eta$}}_{:,j}, \tilde{\mathcal{K}}),
\end{aligned}
\label{indc_cond_prior}
\end{equation}
and, $\mbox{\boldmath{$\eta$}}_{:,j}=\mathcal{K}_{fu}\mathcal{K}_{uu}^{-1}\mathbf{u}_{:,j},~~~\tilde{\mathcal{K}}=\mathcal{K}_{ff}-\mathcal{K}_{fu}\mathcal{K}_{uu}^{-1}\mathcal{K}_{uf}$. Using the inducing points, and by multiplying the distributions given in Eqs.~(\ref{prior_marg_indc}),and (\ref{indc_cond_prior}), the expanded joint probability distribution of $(\mathcal{H}, \mathbb{Y}, \mathcal{F}, \mathcal{U})$ can be expressed as:
\begin{equation}
\begin{aligned}
p(\mathcal{H}, \mathbb{Y}, \mathcal{F}, \mathcal{U}\mid \mbox{\boldmath{$\theta$}}) &= p(\mathbf{Y}\mid \mathcal{F},\sigma^{2})p(\mathcal{F}\mid \mathcal{H}, \mathcal{H}_{u}, \mathcal{U}, \mbox{\boldmath{$\theta$}}_{f})p(\mathcal{U}\mid \mathcal{H}_{u})p(\mathcal{H}) \\
&= p(\mathcal{H})\{\prod_{j=1}^{d}p(\mathbf{y}_{:,j}\mid \mathbf{f}_{:,j})p(\mathbf{f}_{:,j}\mid \mathbf{u}_{:,j},\mathcal{H}, \mathcal{H}_{u} )p(\mathbf{u}_{:,j}\mid \mathcal{H}_{u})\}
\label{joint_dist_augm}
\end{aligned}
\end{equation}
\noindent
Moreover, using the variational inference, the true posterior distribution can be approximated as:
\begin{equation}
\begin{aligned}
q(\mathbb{F}, \mathcal{U}, \mathcal{H}\mid \mathbb{Y}) &= q(\mathbb{F}\mid  \mathcal{U}, \mathcal{H}, \mathbf{Y})q(\mathcal{U}\mid \mathbb{Y})q(\mathcal{H}\mid \mathbb{Y}) \\
&= \left(\prod_{j=1}^{\nu}p(\mathbf{f}_{:,j}\mid \mathbf{u}_{:,j}, \mathcal{H})q(\mathbf{u}_{:,j})\right)q(\mathcal{H}\mid \mathbb{Y}),\label{Var_join_dist}
\end{aligned}
\end{equation}
where, $q(\mathcal{H}\mid \mathbb{Y})=\mathcal{N}(\mathcal{H}\mid \mbox{\boldmath{$\mu$}}, \mathbf{S})$ and $q(\mathcal{U}\mid \mathbb{Y})=\prod_{j=1}^{d}q(\mathbf{u}_{:,j}\mid \mathbb{Y})$ is considered as an arbitrary variational distribution~\cite{Damianou:variational15}. Finally, the variational lower bound can be calculated as follows:
\begin{equation}
\begin{aligned}
\mathcal{L}(q(\mathcal{H}\mid \mathbb{Y}), q(\mathcal{U}\mid \mathbb{Y})) &= \int q(\mathcal{F}, \mathcal{U}, \mathcal{H}\mid \mathbb{Y})\log \frac{p(\mathbb{Y},\mathcal{F}, \mathcal{U}, \mathcal{H})}{q(\mathcal{F}, \mathcal{U}, \mathcal{H}\mid \mathbb{Y})}~d\mathcal{H}d\mathcal{F}d\mathcal{U} \\
& = \hat{\mathcal{L}}(q(\mathcal{H}\mid \mathbb{Y}),q(\mathcal{U}\mid \mathbb{Y}))-KL(q(\mathcal{H}) \parallel p(\mathcal{H})),\label{orig_LOB}
\end{aligned}
\end{equation}
\noindent
The Kullback-Leibler ($KL$) divergence between $q(\mathbf{u}_{:,j}\mid \mathbb{Y})$ and $p(\mathbf{u}_{:,j})$ can be derived as:
\begin{equation}
\begin{aligned}
KL(q(\mathbf{u}_{:,j}\mid \mathbb{Y})\parallel p(\mathbf{u}_{:,j})) &=
\int q(\mathbf{u}_{:,j}\mid \mathbb{Y}) \log \frac{q(\mathbf{u}_{:,j}\mid \mathbb{Y})}{p(\mathbf{u}_{:,j})}d\mathbf{u}_{:,j} \\
&= -[E_{q(\mathbf{u}_{:,j}\mid \mathbb{Y})}(p(\mathbf{u}_{:,j}))+\mathcal{H}_{q(\mathbf{u}_{:,j}\mid \mathbb{Y})}] \label{KL_qp}
\end{aligned}
\end{equation}
where, $\mathcal{H}_{q(\mathbf{u}_{:,j}\mid \mathbb{Y})}$ denotes the entropy of the distribution. Since $\hat{\mathcal{L}}_{j}(q(\mathcal{H}\mid \mathbb{Y}))\geq \hat{\mathcal{L}}_{j}(q(\mathcal{H}\mid \mathbb{Y}), q(\mathbf{u}_{:,j}\mid \mathbf{Y}))$, the lower bound $\hat{\mathcal{L}}_{j}()$ can be analytically computed as follows:
\begin{equation}
\begin{split}
\hat{\mathcal{L}}_{j}(q(\mathcal{H}\mid \mathbb{Y}))&=\log\left[\frac{\sigma^{n}|\mathcal{K}_{uu}|^{0.5}}{(2\pi)^{\frac{n}{2}}|\sigma^{-2}\mbox{\boldmath{$\Phi$}}_{2}+\mathcal{K}_{uu}|^{0.5}}\exp\{-\frac{1}{2}\mathbf{y}_{:,j}^{T}\mathcal{Z}\mathbf{y}_{:,j}\}\right]\\
&-\frac{\phi_{0}}{2\sigma^{2}}+\frac{tr(\mathcal{K}_{uu}^{-1}\mbox{\boldmath{$\Phi$}}_{2})}{2\sigma^{2}},
\end{split}
\label{LOB_com_fin}
\end{equation}
where,
\begin{align}
\begin{split}
\phi_{0}&=tr(E_{q(\mathbf{H}\mid \mathbf{Y})}[\mathbf{K}_{ff}]),~~~\mbox{\boldmath{$\Phi$}}_{1}=E_{q(\mathbf{H}\mid \mathbf{Y})}[\mathbf{K}_{fu}],~~~~\mbox{\boldmath{$\Phi$}}_{2}=E_{q(\mathbf{H}\mid \mathbf{Y})}[\mathbf{K}_{uf}\mathbf{K}_{fu}] \nonumber \\
\mathcal{Z}&=\sigma^{-2}\mathbf{I}_{n}-\sigma^{-4}\mbox{\boldmath{$\Phi$}}_{1}(\sigma^{-2}\mbox{\boldmath{$\Phi$}}_{2}+\mathbf{K}_{uu})^{-1}\mbox{\boldmath{$\Phi$}}_{1}^{T}.\nonumber
\end{split}
\end{align}
\noindent
By substituting the $\hat{\mathcal{L}}_{j}(q(\mathcal{H}\mid \mathbb{Y}))$ into Eq.~(\ref{orig_LOB}), the final representation of variational lower bound for $\log p(\mathbb{Y})$ is given by:
\begin{equation}
\mathcal{L}(q(\mathcal{H}\mid \mathbb{Y}))=\hat{\mathcal{L}}(q(\mathcal{H}\mid \mathbb{Y}))-KL(q(\mathcal{H}\mid \mathbb{Y}) \parallel p(\mathcal{H})),\label{LB_MLofY}
\end{equation}
where,
\begin{equation}
\begin{aligned}
\hat{\mathcal{L}}(q(\mathcal{H}\mid \mathbb{Y})) &=\sum_{j=1}^{d} \hat{\mathcal{L}}_{j}(q(\mathcal{H}\mid \mathbb{Y})), \\
KL(q(\mathcal{H}\mid \mathbb{Y}) \parallel p(\mathcal{H}))&=\frac{1}{2}\sum_{i=1}^{n}tr\left(\mbox{\boldmath{$\mu$}}_{i,:}\mbox{\boldmath{$\mu$}}_{i,:}^{T}+\mathcal{S}_{i}-\log \mathcal{S}_{i}\right)-\frac{nq}{2}, \\
q(\mathcal{H}\mid \mathbb{Y})&=\mathcal{N}(\mathcal{H}\mid \mbox{\boldmath{$\mu$}}, \mathcal{S})=\prod_{i=1}^{n}\mathcal{N}(\mathcal{h}_{i,:}\mid \mbox{\boldmath{$\mu$}}_{i,:},\mathcal{S}_{i}),
\end{aligned}\label{GPLVM_QH}
\end{equation}
and each covariance matrix $\mathcal{S}_{i}$ is diagonal. Hitherto, the Bayesian GP-LVM has been introduced. The variational parameters of $q(\textbf{h})$ and kernel parameters have been approximated by maximizing the lower bound of the $p(\mathbb{Y})$ as given in Eq.~(\eqref{LB_MLofY}). Moreover, the variational posterior distribution, $q(\textbf{h})$ has been approximated as given in Eq.~(\eqref{GPLVM_QH}). As mentioned earlier, the Bayesian application is required to be implemented along with the corresponding variational posterior distributions. This is essential in order to generate the data to build the deep GP model as a generative model. The main steps required to train a deep GP model with $L$ hidden layers and subsequently to construct the deep GP model are as follows:
\begin{enumerate}
	\item Given the observed output data $\mathbb{Y}$, apply Bayesian GP-LVM in order to ($i$) generate the $G^{th}$ latent variable, $\textbf{h}_{G}$ with dimension $q_{G}$; ($ii$) approximate the variational posterior distribution, $q(\textbf{h}_{G})$; and ($iii$) provide the generated data, $\mathcal{H}_{G}\in \mathbb{R}^{n\times q_{G}}$ required to generate the other latent variables.
	
	\item Given the generated data, $\mathcal{H}_{G}$ apply Bayesian GP-LVM in order to ($i$) generate the $(G-1)^{th}$ latent variable, $\textbf{h}_{G-1}$ with dimension $q_{G-1}$; ($ii$) approximate the corresponding variational posterior distribution, $q(\textbf{h}_{G-1})$; and ($iii$) provide the generated data, $\mathcal{H}_{G-1}\in \mathbb{R}^{n\times q_{G-1}}$ required to generate the other latent variables.
	
	\item Continue this process $G$ times to generate $G$ latent variables $(\textbf{h}_{1},\ldots,\textbf{h}_{G})$ and their corresponding posterior distributions $\{q(\textbf{h}_{i})\}_{i=1}^{G}$.
	
	\item Approximate the variational parameters and model parameters including the correlation functions parameters by optimizing the lower bound of $p(\textbf{y}\mid \textbf{x})$ as given in Equation \eqref{FLB}.
	\item Once the deep GP model is trained and the latent variables with reduced dimensionality are generated, we can use the trained deep GP model given in Eq.~\eqref{nested_GP} for prediction and UQ tasks (including the computation of QoI and propagate the uncertainty from inputs to outputs and across the hierarchy in a tractable way) as discussed in Subsection~\ref{Sec:Pre_MC_GPLVM}.   
\end{enumerate}

\subsection{Deep Gaussian Processes for Learning High-dimensional data}\label{sec:DeepGP}

Bayesian GP-LVM has been introduced in the previous section as an effective approach to provide probabilistic modelling of high dimensional data that lies on a nonlinear latent space. However, this model is suitable for unsupervised learning only and therefore it cannot be directly implemented for regression modelling or supervised learning. In this section, we introduce a hierarchical GP model which benefits from the nice properties of the Bayesian GP-LVM to reduce the dimensionality of the given data and can be also efficiently used for regression modelling. This model is called Deep GP and is a deep belief network based on GP mappings~\cite{Damianou.2015}. A deep GP model consists of $G$ latent/hidden layers as demonstrated in Fig.~\ref{DGP-sup}.
\begin{figure}[htb]
	\begin{center}
		\scalebox{1}{\includegraphics{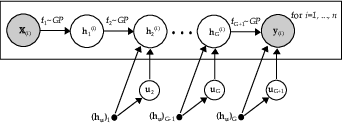}}
	\end{center}
	\caption{A graphical model representation of the supervised deep GP structure inducing variables and $G$ hidden layers.\label{DGP-sup}}
\end{figure}
\noindent
For the deep GP model developed for supervised learning (or regression scenarios), the inputs, $\textbf{x}\in \mathbb{R}^{\kappa}$ are placed on the parent nodes, the outputs, $\textbf{y}\in \mathcal{R}^{\nu}$ are placed in the leaves of the hierarchy, and the intermediate nodes are used to present the latent variables $\textbf{h}_{i}\in \mathbb{R}^{q_{i}},~~i=1,\ldots,G$.
These $G$ hidden layers are hierarchically connected to each other such that the $i^{th}$ latent variable, $\textbf{h}_{i} \in \mathbb{R}^{q_{i}}$ constitutes the output of that layer and acts as an input for the generative procedure of the subsequent layer, $i+1$ (see Equation~\eqref{nested_GP}).
In this section, we adopt the Bayesian variational framework introduced above to train the deep GP model by marginalising out the latent variables and optimizing the variational lower bound of the marginal likelihood of $p(\textbf{y}\mid \textbf{x})$. In this regard, the joint distribution of a deep GP model with $G$ hidden layers can be given by
\begin{equation}
p(\textbf{y}, \{\textbf{h}_{i}\}_{i=1}^{G}\mid \textbf{x})=p(\textbf{y}\mid \textbf{h}_{G})\prod_{i=2}^{L}p(\textbf{h}_{i}\mid \textbf{h}_{i-1})p(\textbf{h}_{1} \mid \textbf{x}),\label{prob-deepGp}
\end{equation}
where
\begin{equation}
\begin{aligned}
p(\textbf{h}_{1} \mid \textbf{x})&= \int p(\textbf{h}_{1} \mid \textbf{f}_{1})p(\textbf{f}_{1} \mid \textbf{x})d\textbf{f}_{1},\\
p(\textbf{h}_{i} \mid \textbf{h}_{i-1})&= \int p(\textbf{h}_{i} \mid \textbf{f}_{i})p(\textbf{f}_{i} \mid \textbf{h}_{i-1})d\textbf{f}_{i},~ \text{for}~i=2,\ldots,G. 
\end{aligned}\label{cond-prob1}
\end{equation}
\noindent
The first and second terms of the integrals above can be respectively given by:
\begin{align}
p(\textbf{f}_{1} \mid \textbf{x})=\mathcal{N}(\textbf{f}_{1}\mid \textbf{0}, \mathcal{K}_{\textbf{f}_{1}\textbf{f}_{1}}),~~p(\textbf{h}_{i} \mid \textbf{f}_{i})=\mathcal{N}(\textbf{h}_{i}\mid \textbf{f}_{i}, \sigma^{2}_{i}\textbf{I}),\nonumber \\
p(\textbf{f}_{i} \mid \textbf{h}_{i-1})=\mathcal{N}(\textbf{f}_{i}\mid \textbf{0}, \mathcal{K}_{\textbf{f}_{i}\textbf{f}_{i}}),~~ p(\textbf{y} \mid \textbf{h}_{G})=\mathcal{N}(\textbf{y}\mid \textbf{f}_{G}, \sigma^{2}_{G}\textbf{I}).\nonumber
\end{align}
For a better understanding of the computational intractability of learning deep GP model, a simpler version of this model with two hidden layers is considered as shown in Fig.~\ref{DGP-2L}.} 
\begin{figure}[htb]
\begin{center}
	\scalebox{1}{\includegraphics{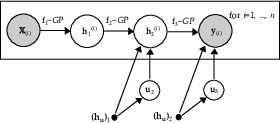}}
\end{center}
\caption{A graphical model representation of the supervised deep GP structure with inducing variables and $G=2$ hidden layers.\label{DGP-2L}}
\end{figure}
The marginal likelihood of $\textbf{y}$ given $\textbf{x}$ of this deep GP model consists of 2 hidden layers is given as follows:
\[
p(\textbf{y} \mid \textbf{x})=\int p(\textbf{y} \mid \textbf{h}_{2}) \int p(\textbf{h}_{2} \mid \textbf{h}_{1}) p(\textbf{h}_{1} \mid \textbf{x})~d\textbf{h}_{1}d\textbf{h}_{2}.
\]
In the nested integral above, computing $p(\textbf{h}_{1} \mid \textbf{x})$ is analytically feasible, since $\mathbf{x}$ is observed (and does not need to be probabilistically generated by applying GP-LVM on $\mathbf{h}_{1}$) and propagating it through the nonlinear mapping $\mathbf{f}_{1}$ is tractable. Therefore, $p(\textbf{h}_{1} \mid \textbf{x})$ can be analytically computed by computing the following Gaussian integral:
\begin{equation}
p(\textbf{h}_{1} \mid \textbf{x})=\int p(\textbf{h}_{1} \mid \textbf{f}_{1})p(\textbf{f}_{1} \mid \textbf{x})d\textbf{f}_{1}=\mathcal{N}(\mathbf{h}_{1}|\mathbf{0},\mathcal{K}_{\textbf{f}_{1}\textbf{f}_{1}}+\sigma^{2}_{1}\textbf{I})\label{eq_Prob_h1gx}
\end{equation}
}
Given the observed input $\textbf{x}$, the nested integral can be written as:
\[
p(\textbf{h}_{2} \mid \textbf{x})=\int\int p(\textbf{h}_{2}\mid \textbf{f}_{2})p(\textbf{f}_{2}\mid \textbf{h}_{1}) p(\textbf{h}_{1} \mid \textbf{x})d\textbf{h}_{1}d\textbf{f}_{2}=E^{p(\textbf{h}_{1} \mid \textbf{x})}[p(\textbf{h}_{2} \mid \textbf{h}_{1})].
\]
Computing this integral is intractable, as $p(\textbf{h}_{1} \mid \textbf{x})$ cannot be propagated through the function $p(\textbf{f}_{2}\mid \textbf{h}_{1})$ which is constructed using a nonlinear covariance function, {$\mathcal{K}_{\textbf{f}_{1}\textbf{f}_{1}}$}. Thus, there is a similar complexity in every layer to compute $\tilde{p}({\textbf{h}_{i}})=E^{\tilde{p}(\textbf{h}_{i-1})}[p(\textbf{h}_{i} \mid \textbf{h}_{i-1})]$, {for $i=2,\ldots,G$}. This computational intractability can be overcome by extending the Bayesian variational method introduced previously. In this regard, the probability space of the layers with the inducing (or auxiliary) variables {$(\textbf{h}_{u})_{i-1}$ and $\textbf{u}_{i}$ augmented initially for $i=2,\ldots,G+1$}, as shown in Fig \ref{DGP-sup}. The conditional probability, $p(\textbf{h}_{i} \mid \textbf{h}_{i-1})$ is then modified by including these inducing variables as $p(\textbf{h}_{i} \mid \textbf{h}_{i-1}, \textbf{u}_{i})$, where the values of $\textbf{u}_{i}=\left[u_{i}^{(1)}, u_{i}^{(2)}, \ldots, u_{i}^{(m_{i})}\right]$ are evaluated at the inducing input points, 
\[
{(\textbf{h}_{u})_{i-1}=\left[(h_{u})_{i-1}^{(1)}, \ldots, (h_{u})_{i-1}^{(m_{i})}\right],~~\textrm{for}~~i=2,\ldots,G+1.}
\] 
Suppose, generating the hidden layer $\mathbf{h}_{2}$ given the observed data,\\ $\mathbb{Y}=(\mathbf{y}_{1},\ldots, \mathbf{y}_{n})$ using the GP-LVM. In order to tackle the computational complexity we augment the corresponding latent function $\mathbf{f}_{3}$ (or joint distribution of $(\mathbf{y}, \mathbf{f}_{3}, \mathbf{h}_{2})$) with the inducing variables $\mathbf{u}_{3}$. Similarly, we augment the latent function $\mathbf{f}_{2}$ with the inducing variables $\mathbf{u}_{2}$ to overcome the computational intractability of generating the hidden layer $\mathbf{h}_{1}$. I should be noted that the function $\mathbf{f}_{1}$ with GP prior that maps $\mathbf{x}$ to $\mathbf{h}_{1}$ is not considered as the latent function and therefore it does not need to be augmented by the inducing points. This is mainly because the computation of $p(\mathbf{h}_{1}\mid \mathbf{x})$ as easily derived in Equation (\ref{eq_Prob_h1gx}) is tractable. In other words, it is only required to augment the latent functions $\mathbf{f}_{i}$.
Therefore, the conditional probability defined in Eq.~(\ref{cond-prob1}) can now be expressed as:
\begin{equation}
p(\textbf{h}_{i} \mid \textbf{h}_{i-1}, \textbf{u}_{i})=\int p(\textbf{h}_{i} \mid \textbf{f}_{i})p(\textbf{f}_{i} \mid \textbf{h}_{i-1},\textbf{u}_{i})d\textbf{f}_{i},\label{cond-prob}
\end{equation}
where,
\begin{equation}
\begin{aligned}
p(\textbf{h}_{i} \mid \textbf{f}_{i}) &= \mathcal{N}(\textbf{h}_{i}\mid \textbf{f}_{i}, \sigma^{2}_{i}\textbf{I})~\text{and},~
p(\textbf{f}_{i} \mid \textbf{h}_{i-1},\textbf{u}_{i})= \mathcal{N}(\textbf{f}_{i}\mid \mbox{\boldmath{$\mu$}}_{i}, \tilde{\mathcal{K}}_{i}), \\
\mbox{\boldmath{$\mu$}}_{i} &=\mathcal{K}_{\textbf{f}_{i-1}\textbf{u}_{i-1}}\mathcal{K}^{-1}_{\textbf{u}_{i-1}\textbf{u}_{i-1}}\textbf{u}_{i}, \\
\tilde{\mathcal{K}}_{i}&=\mathcal{K}_{\textbf{f}_{i-1}\textbf{f}_{i-1}}-\mathcal{K}_{\textbf{f}_{i-1}\textbf{u}_{i-1}}\mathcal{K}^{-1}_{\textbf{u}_{i-1}\textbf{u}_{i-1}}\mathcal{K}_{\textbf{u}_{i-1}\textbf{f}_{i-1}}.
\end{aligned}
\end{equation}
\noindent
The computation of $p(\textbf{f}_{i} \mid \textbf{h}_{i-1},\textbf{u}_{i})$ is intractable. Thus, similar to the previous section for variational GP-LVM ~\cite{DBLP:journals/jmlr/TitsiasL10}), the bound can be computed as:
\begin{equation}
\begin{split}
\log p(\textbf{y}, \{\textbf{h}_{i}\}_{i=1}^{G}\mid \{\textbf{u}_{i}\}_{i=1}^{G+1}, \textbf{x}) &=
\log p(\textbf{y}\mid \textbf{h}_{G},\textbf{u}_{G+1})+\sum_{i=2}^{G} \log p(\textbf{h}_{i}\mid \textbf{h}_{i-1},\textbf{u}_{i}) \\
&+\log p(\textbf{h}_{1}\mid \textbf{x})\nonumber 
\geq \mathfrak{L}=\log p(\textbf{h}_{1}\mid \textbf{x})+ \sum_{i=2}^{G+1} \mathfrak{L}_{i},\label{bound1}
\end{split}
\end{equation}
where, $\mathfrak{L}_{G+1}$ is the lower bound for $\log p(\textbf{y}\mid \textbf{h}_{G},\textbf{u}_{G+1})$, and $\mathfrak{L}_{i}$'s are respectively lower bounds for $\log p(\textbf{h}_{i}\mid \textbf{h}_{i-1},\textbf{u}_{i})$. The lower bound terms, $\mathfrak{L}_{i}$ can be analytically derived~\cite{journals/jmlr/Titsias09} as follows:
\begin{equation}
\mathfrak{L}_{i}=\log \mathcal{N}(\textbf{h}_{i}\mid \mbox{\boldmath{$\mu$}}_{i}, \sigma^{2}_{i}\textbf{I})-\frac{1}{2\sigma^{2}_{i}}tr(\tilde{\mathcal{K}}_{i}).\label{LB1}
\end{equation}
\noindent
Hitherto, a lower bound for, $\log p(\textbf{y}\mid \textbf{x})$ has been computed by further integrating out $\textbf{h}_{i}$ and $\textbf{u}_{i}$. To variationally approximate the following term
\begin{equation}
\log p(\textbf{y}\mid \textbf{x})= \log \int p(\textbf{y}, \{\textbf{h}_{i}\}_{i=1}^{G}\mid \{\textbf{u}_{i}\}_{i=1}^{G+1},\textbf{x})\prod_{i=1}^{G} d\textbf{h}_{i}\prod_{i=1}^{G+1} p(\textbf{u}_{i})d\textbf{u}_{i}.
\end{equation}
It is required to introduce the following variational distribution, as thus:
\begin{equation}
\mathcal{L}_{1}=\prod_{i=1}^{G}q(\textbf{u}_{i+1})q(\textbf{h}_{i})\label{VB_uh}.
\end{equation}
Given this distribution and using the Jensen's inequality, the bound $\mathcal{L}\leq p(\textbf{y}\mid \textbf{x})$ can be given by:
\begin{equation}
\begin{split}
\mathcal{L}&=\int \mathcal{L}_{1} \log \frac{p(\textbf{y}, \{\textbf{h}_{i}\}_{i=1}^{G}\mid \{\textbf{u}_{i}\}_{i=1}^{G1},\textbf{x})\prod_{i=1}^{G+1}p(\textbf{u}_{i})}{\mathcal{L}_{1}}d\{\textbf{h}_{i}\}_{i=1}^{G}d\{\textbf{u}_{i}\}_{i=1}^{G} \nonumber \\
&=E^{\mathcal{L}_{1}}\left[\log p(\textbf{y}, \{\textbf{h}_{i}\}_{i=1}^{G}\mid \{\textbf{u}_{i}\}_{i=1}^{G+1},\textbf{x})\right]-\sum_{i=1}^{G+1}KL(q(\textbf{u}_{i}) \|  p(\textbf{u}_{i}))+\sum_{i=1}^{G} \mathcal{H}(q(\textbf{h}_{i})),
\end{split}\label{LBv1}
\end{equation}
where, $\mathcal{H}(q(\textbf{h}_{i}))$ denotes the entropy of $q(\textbf{h}_{i})$, $q(\textbf{u}_{i})$ and $q(\textbf{h}_{i})$ denote the variational distributions which are given by:
\begin{align}
q(\textbf{h}_{i})=\mathcal{N}(\textbf{h}_{i}\mid \textbf{m}_{i}, \mathcal{S}_{i}), \nonumber \\
q(\textbf{u}_{i})=\mathcal{N}(\textbf{u}_{i}\mid \mbox{\boldmath{$\nu$}}_{i}, \mbox{\boldmath{$\Sigma$}}_{i}).\nonumber
\end{align}
By using the Gaussian distributions for  $q(\textbf{u}_{i})$ and $q(\textbf{h}_{i})$, all the terms in Equation\eqref{LB1} are tractable and therefore Equation \eqref{VB_uh} can be rewritten as follows:
\begin{equation}
\begin{aligned}
\mathcal{L}_{1} &=\prod_{i=1}^{G}\mathcal{N}(\textbf{h}_{i}\mid \textbf{m}_{i}, \mathcal{S}_{i})\mathcal{N}(\textbf{u}_{i+1}\mid \mbox{\boldmath{$\nu$}}_{i+1}, \mbox{\boldmath{$\Sigma$}}_{i+1})\nonumber \\
&=\prod_{i=1}^{G}\left(\mathcal{N}(\textbf{u}_{i+1}\mid \mbox{\boldmath{$\nu$}}_{i+1}, \mbox{\boldmath{$\Sigma$}}_{i+1})\prod_{j=1}^{n}\mathcal{N}({h}_{i}^{(j)}\mid {m}_{i}^{(j)}, {s}_{i}^{(j)})\right). 
\end{aligned}\label{VB_uhv2_1}
\end{equation}
\noindent
If the dimension of each latent layer, $q_{i}>1$, the variational distribution will take the following form
\begin{equation}
{\mathcal{L}_{1}=q(\mathcal{H}_{i})q(\mathcal{U}_{i})}\label{VB_uhv2}
\end{equation}
{where, the variational distribution associated with the latent space distribution can be factorise as:
\begin{equation}
\begin{aligned}
q(\mathcal{H}_{i})&=\prod_{j=1}^{q_{1}}\mathcal{N}(\mathbf{h}_{1}^{(j)}| \mathbf{m}_{1}^{(j)}, \mathcal{S}_{1}^{(j)})\prod_{i=2}^{L}\prod_{k=1}^{n}\mathcal{N}(\mathbf{h}_{i}^{(k,:)}\mathbf{m}_{i}^{(k,:)}, \mathcal{S}_{i}^{(k,:)}), \\
q(\mathcal{U}_{i})&=\prod_{i=1}^{G}\prod_{j=1}^{q_{i}+1}\mathcal{N}(\textbf{u}_{i+1}^{(j)}\mid \mbox{\boldmath{$\nu$}}_{i+1}^{(j)}, \mbox{\boldmath{$\Sigma$}}_{i+1}^{(j)}), 
\end{aligned}
\end{equation}
\noindent
where, ${\mathcal{S}}_{1}^{(j)}$ is a full $n\times n$ matrix,} ${\mathcal{S}}_{i}^{(k)}$ are diagonal $q_{i}\times q_{i}$ matrices, and $\mbox{\boldmath{$\Sigma$}}_{i}^{(j)}$ are full $m_{i}\times m_{i}$ matrices.
By using the expression for $\mathcal{L}_{1}$ and the bound given in Equation~\eqref{bound1}, the first term in Equation \eqref{LBv1} can be computed as follows:
\begin{equation}
\begin{split}
E^{\mathcal{L}_{1}}\left[\log p(\textbf{y}, \{\textbf{h}_{i}\}_{i=1}^{G}\mid \{\textbf{u}_{i}\}_{i=1}^{G+1},\textbf{x})\right]& =
E^{q(\textbf{h}_{1})}\left[\log p(\textbf{h}_{1}\mid \textbf{x})\right]\\
&+ \sum_{i=2}^{G+1} E^{q(\textbf{h}_{i-1})q(\textbf{h}_{i})q(\textbf{u}_{i})}\left[\mathfrak{L}_{i}\right].
\end{split}\label{LBv2}
\end{equation}
\noindent
By substituting Equation \eqref{LBv2} into Equation \eqref{LBv1}, the final form of the bound $\mathcal{L}\leq \log p(\textbf{y}\mid \textbf{x})$ can be presented as thus:
\begin{equation}
\begin{aligned}
\mathcal{L} &= \sum_{i=2}^{G+1}\left(E^{q(\textbf{h}_{i-1})q(\textbf{h}_{i})q(\textbf{u}_{i})}\left[ \log \mathcal{N}(\textbf{h}_{i} \mid  \mbox{\boldmath{$\mu$}}_{i}, \sigma^{2}_{i}\textbf{I})\right]-\frac{1}{2\sigma^{2}_{i}}E^{q(\textbf{h}_{i-1})}\left[tr(\tilde{\mathcal{K}}_{i})\right]\right) \\
&-KL(q(\textbf{h}_{1}) \|  p(\textbf{h}_{1}\mid \textbf{x}))-\sum_{i=2}^{G+2} KL(q(\textbf{u}_{i}) \|  p(\textbf{u}_{i}))+\sum_{i=2}^{G} \mathcal{H}(q(\textbf{h}_{i})),
\end{aligned}\label{FLB}
\end{equation}
\noindent
It is worth noting that all the terms of the bound presented above are tractable. In particular, the $KL$ and entropy terms are trivial to evaluate and can be simply derived as follows:
\[
\mathcal{H}(q(\textbf{h}_{i}))=\frac{nq_{i}}{2}(1+\log(2\pi))+\frac{1}{2}\sum_{k=1}^{n}\sum_{j=1}^{q_{i}}\log(\mathcal{S}_{i}^{(k)})^{(j,j)},
\]
where $(\mathcal{S}_{i}^{(k)})^{(j,j)}$ represents the $j^{th}$ component in the diagonal of the covariance matrix $\mathcal{S}_{i}^{(k)}$. The first term given in Equation~(\ref{FLB}) can be obtained by expanding the Gaussian forms. It is trivial that the expectations with respect to the $q(\textbf{h})$ terms turns the covariance matrices into $\{\phi_{0}, \mbox{\boldmath{$\Phi$}}_{1}, \mbox{\boldmath{$\Phi$}}_{2}\}$ statistics for every layer. Finally, the final form of variational lower bound can be optimised with respect to the model parameters ($\{\sigma_{i}^{2}, \mbox{\boldmath{$\theta$}}_{i}\}$) and the variational parameters ($\{\textbf{u}_{i}, \textbf{m}_{i}, \mathcal{S}_{i}, \mbox{\boldmath{$\nu$}}_{i}, \mbox{\boldmath{$\Sigma$}}_{i}\}$) using a gradient based optimisation algorithm~\cite{Damianou.2015}. The steps required to train a deep GP with $G$ hidden layers given the observed data are thus:
\begin{enumerate}
\item Let ($i$) The observed input $\mathbb{X}\in \mathbb{R}^{n \times \kappa}$ and output $\mathbb{Y}\in \mathbb{R}^{n\times \nu}$; ($ii$) $G$, the pre-fixed number of hidden layers; ($iii$) $q_{i}$, the pre-fixed dimensionality of each hidden layer for $i=1,\ldots,G$; ($iv$) $m_{i}$, the inducing variables on the $i^{th}$ hidden layer; and ($v$) $T$, the number of training iterations, 

\item Initialise $\textbf{h}_{G}=\mathrm{GPLVM}(\textbf{Y}, q_{L})$ and $\textbf{h}_{i-1}=\mathrm{GPLVM}(\textbf{h}_{i}, q_{i}),~i=2,\ldots,G$ (or by other DM approaches, e.g, PPCA, PCA, isomap);\\ and $\{\textbf{m}_{i}, \mathcal{S}_{i}, \mbox{\boldmath{$\nu$}}_{i}, \mbox{\boldmath{$\Sigma$}}_{i}, \textbf{u}_{i}, \sigma^{2}_{i}, \mbox{\boldmath{$\omega$}}_{i}\}$

\item Augment each hidden layer, $\textbf{h}_{i}$ and $\textbf{y}$ with $m_{i}$ inducing variables, $\{\textbf{u}_{i}\}_{2}^{G+1}$

\item For $T$ iterations, optimize the lower bound for $\log p(\mathbb{Y}\mid \mathbb{X})$ given in Equation.~(\ref{FLB}) w.r.t $\{\textbf{m}_{i}, \mathcal{S}_{i}, \mbox{\boldmath{$\nu$}}_{i}, \mbox{\boldmath{$\Sigma$}}_{i}, \textbf{u}_{i}, \sigma^{2}_{i}, \mbox{\boldmath{$\omega$}}_{i}\}$

\item Continue the iterations until the stopping rule is reached, then obtain the optimised $\mathcal{L}$,  $\{\hat{\textbf{m}}_{i}, \hat{\mathcal{S}}_{i}, \widehat{\mbox{\boldmath{$\nu$}}}_{i}, \widehat{\mbox{\boldmath{$\Sigma$}}}_{i}, \widehat{\textbf{u}}_{i}, \widehat{\sigma^{2}}_{i}, \widehat{\mbox{\boldmath{$\omega$}}}_{i}\}$.
\end{enumerate}

\subsection{Predictions by the Deep GP Model}\label{Sec:Pre_MC_GPLVM}

In this section, we use the trained deep GP model along with a vector of test input points, $\textbf{x}^{*}$ to predict the corresponding outputs, $\textbf{y}^{*}$. Initially, it is required to evaluate, $p(\textbf{h}_{1}^{*}\mid \textbf{h}_{1}, \mathbb{X},\textbf{x}^{*})$ using a GP model with the following mean $\mbox{\boldmath{$\mu$}}_{1}^{*}$ and $\mbox{\boldmath{$\Sigma$}}_{1}^{*}$, as thus:
\begin{equation}
\begin{aligned}
\mbox{\boldmath{$\mu$}}_{1}^{*} &=k(\textbf{x}^{*}, \textbf{X})^{T}(k(\textbf{X}, \textbf{X})+\hat{\sigma^{2}}_{1}I)^{-1}\textbf{h}_{1}, \\
\hat{\mbox{\boldmath{$\Sigma$}}}_{1}^{*} &=k(\textbf{x}^{*}, \textbf{x}^{*})-k(\textbf{x}^{*}, \textbf{X})^{T}(k(\textbf{X}, \textbf{X})+\hat{\sigma^{2}}_{1}I)^{-1}k(\textbf{x}^{*}, \textbf{X}).\label{PR_mean1Var1}
\end{aligned}
\end{equation}
\noindent
Afterwards,the predictive mean and covariance given in Equation~(\eqref{PR_mean1Var1}) can be computed through the following steps \cite{rasmussen:williams:2006}:
\begin{enumerate}
	\item Let, $\mathbb{X}$ (inputs), $\textbf{h}_{1}\in \mathbb{R}^{n\times q_{1}}$ (trained hidden values), $\textbf{x}^{*}\in \mathbb{R}^{n^{*}\times p}$ (test inputs); and optimized variational and model hyper-parameters
	
	\item Evaluate $G_{h_{1}}=cholesky(k(\mathbb{X}, \mathbb{X})+\hat{\sigma^{2}}_{1}I)$ and $\alpha_{h_{1}}=G_{h_{1}}^{T}\setminus(G_{h_{1}}\setminus \textbf{h}_{1})$
	in order to calculate the predictive mean as ${\mbox{\boldmath{$\Sigma$}}}_{1}^{*}=k(\textbf{x}^{*}, \textbf{x}^{*})-v_{h_{1}}^{T}v_{h_{1}}$
	
	\item Evaluate $v_{h_{1}}=L_{h_{1}}\setminus k(\textbf{x}^{*}, \textbf{X})$ in order to calculate the predictive variance as ${\mbox{\boldmath{$\Sigma$}}}_{1}^{*}=k(\textbf{x}^{*}, \textbf{x}^{*})-v_{h_{1}}^{T}v_{h_{1}}$
\end{enumerate}

The predictive mean of the test points given in Equation (\ref{PR_mean1Var1}) can be used as the predictions of the latent variable values on the first layer given $\textbf{x}^{*}$. These values can be considered as the input test points for the next layer to derive $p(\textbf{h}_{2}^{*}\mid \textbf{h}_{2}, \textbf{h}_{1},\textbf{h}_{1}^{*})$. This prediction process can be similarly continued to the bottom layer to approximate $p(\textbf{y}^{*}\mid \textbf{x}^{*}, \mathbb{X}, \mathbb{Y})$ and predict the values of $\textbf{y}^{*}$ given the selected test input $\textbf{x}^{*}$. Subsequently, by using the GP-LVM models  which is derived during the training process of the deep GP model, the predictive density, $p(\textbf{y}^{*}\mid\textbf{Y}, \textbf{h}_{G}, \textbf{h}_{G}^{*})$ can be approximated. Note that $\textbf{y}^{*}\in \mathbb{R}^{n_{*}\times d}$ denotes the outputs to be predicted, given the test inputs $\textbf{h}_{G}^{*}$ which are derived from $p(\textbf{h}_{G}^{*}\mid \textbf{h}_{G}, \textbf{h}_{G-1}^{*},\textbf{h}_{G-1})$; where $\textbf{h}_{G}$ is the matrix of the $G^{th}$ latent variable values derived from the trained GP-LVM, and $\mathbb{Y})$ is the observed matrix output. Here, we omit the conditioning on $\textbf{h}_{G}^{*},\textbf{h}_{G}$ for clarity of notation. Therefore, the predictive density can be defined as
\begin{equation}
p(\textbf{y}^{*}\mid \textbf{y})=\int{p(\textbf{y}^{*}\mid \textbf{f}_{G}^{*})p(\textbf{f}_{G}^{*}\mid \textbf{h}_{G}^{*}, \textbf{y})p(\textbf{h}_{G}^{*}\mid \textbf{y})d\textbf{f}_{*}d\textbf{h}_{G}^{*}},\label{Pre_ySt_2}
\end{equation}
where, $\textbf{h}_{G}^{*}\in \mathbb{R}^{n_{*}\times q_{G}}$ is the reduced subspace of $\textbf{y}^{*}$, and $\textbf{f}_{G}^{*}\in \mathbb{R}^{n^{*}\times \nu}$ (known as the noisy-free version of $\textbf{y}^{*}$).
Moreover, the term $p(\textbf{f}_{G}^{*}\mid \textbf{h}_{G}^{*}, \textbf{y})$ is approximated by the following variational distribution:
\begin{equation}
q(\textbf{f}_{G}^{*}\mid \textbf{h}_{G}^{*})=\prod_{j=1}^{\nu}q(\textbf{f}_{G}^{*}(:,j)\mid \textbf{h}_{G}^{*})
\end{equation}
where, $q(\textbf{f}_{G}^{*}(:,j)\mid \textbf{h}_{G}^{*})$follows a Gaussian distribution that can be computed analytically~\cite{Damianou.etal.2011}. The third term of Equation~\eqref{Pre_ySt_2}, $p(\textbf{h}_{G}^{*}\mid \textbf{y})$ can be approximated by the following Gaussian variational distribution:
\begin{equation}
q(\textbf{h}_{G}^{*})=\prod_{i=1}^{q_{G}}q(\textbf{h}_{G}^{*}(:,i))=\prod_{i=1}^{q_{G}}\int{p(\textbf{h}_{G}^{*}(:,i)\mid \textbf{h}_{G}(:,i))q(\textbf{h}_{G}(:,i))d\textbf{h}_{G}(:,i)},
\end{equation}
where, $p(\textbf{h}_{G}^{*}(:,i)\mid \textbf{h}_{G}(:,i))$ and $q(\textbf{h}_{G}(:,i))$ are both Gaussian distributions, and $q(\textbf{h}_{G}^{*})$ is also distributed as a Gaussian distribution with the following mean and variance:
\begin{equation}
\begin{aligned}
\mbox{\boldmath{$\mu$}}_{\textbf{h}_{G}^{*}}^{(i)}&=k(\textbf{h}_{G},\textbf{h}_{G}^{*})\bar{\mbox{\boldmath{$\mu$}}}_{G}^{(i)},\nonumber\\
\Sigma_{\textbf{h}_{G}^{*}}^{(i)}&=k(\textbf{h}_{G}^{*},\textbf{h}_{G}^{*})-k(\textbf{h}_{G},\textbf{h}_{G})(k(\textbf{h}_{G}^{*},\textbf{h}_{G}^{*})+(\mbox{\boldmath{$\Lambda$}}_{G}^{(i)})^{-1})k(\textbf{h}_{G}^{*},\textbf{h}_{G}),\nonumber
\end{aligned}
\end{equation}
\noindent
where, $\bar{\mbox{\boldmath{$\mu$}}}_{L}^{(i)}$ and $(\mbox{\boldmath{$\Lambda$}}_{L}^{(i)})^{-1}$ are derived the same way as explained in~\ref{ap:DerThetaAndVarPar}.
\par
As a result, the predictive distribution defined in Equation~(\ref{Pre_ySt_2}) can be approximated as follows:
\begin{equation}
p(\textbf{y}^{*}\mid \textbf{y})=\int{p(\textbf{y}^{*}\mid \textbf{f}_{G}^{*})E^{q(\textbf{h}_{G}^{*})}\left[q(\textbf{f}_{G}^{*}\mid \textbf{h}_{G}^{*})\right]d\textbf{f}_{*}}.\label{Pre_ySt_3}
\end{equation}
The above integral cannot be analytically computed. However, its mean and covariance can be analytically evaluated as:
\begin{equation}
\begin{aligned}
E^{q(\textbf{h}_{G}^{*})}\left(\textbf{f}_{G}^{*}\right)&=\mathcal{V}^{T}\mbox{\boldmath{$\Psi$}}_{1}^{*},\\
Cov^{q(\textbf{h}_{G}^{*})}\left(\textbf{f}_{G}^{*}\right)&=\mathcal{V}^{T}(\mbox{\boldmath{$\Psi$}}_{2}^{*}-\mbox{\boldmath{$\Psi$}}_{1}^{*}(\mbox{\boldmath{$\Psi$}}_{1}^{*})^{T})\mathcal{V}^{T}+\mbox{\boldmath{$\Psi$}}_{0}^{*}I \\
& \quad-tr\left[\left(\mathcal{K}_{mm}^{-1}-(\mathcal{K}_{mm}+\sigma_{L}^{-2}\mbox{\boldmath{$\Phi$}}_{2}^{L})^{-1}\right)\mbox{\boldmath{$\Psi$}}_{2}^{*}\right]I,
\end{aligned}
\end{equation}
\noindent
where,
\begin{equation}
\begin{aligned}
\mathcal{V}&=\sigma_{G}^{-2}(\mathcal{K}_{mm}+\mbox{\boldmath{$\Phi$}}_{2}^{G})^{-1}(\mbox{\boldmath{$\Phi$}}_{1}^{G})^{T}\textbf{y},\\
\mbox{\boldmath{$\Psi$}}_{0}^{*}&=E^{q(\textbf{h}_{G}^{*})}\left(k(\textbf{h}_{G}^{*}, \textbf{h}_{G}^{*})\right), \\
\mbox{\boldmath{$\Psi$}}_{1}^{*}&=E^{q(\textbf{h}_{L}^{*})}\left(K_{m*}\right), \\\mbox{\boldmath{$\Psi$}}_{2}^{*}&=E^{q(\textbf{h}_{L}^{*})}\left(\mathcal{K}_{m*}\mathcal{K}_{*m}\right), 
\end{aligned}\nonumber
\end{equation}
\noindent
and $K_{m*}$ denotes the cross-covariance matrix between $\textbf{h}_{G}$ and $\textbf{h}_{G}^{*}$. The $\mbox{\boldmath{$\Phi$}}^{G}$ statistics can be analytically calculated as explained in Section~\ref{Sec:GP-LVM} and~\ref{ap:Phi}.

Since, $\textbf{y}^{*}$ is a corrupted version of $\textbf{f}_{G}^{*}$, the mean and covariance of $\textbf{y}_{*}$ can be given by:
\begin{equation}
\begin{aligned}
E^{q(\textbf{h}_{G}^{*})}\left(\textbf{y}^{*}\right)&=E^{q(\textbf{h}_{G}^{*})}\left(\textbf{f}_{G}^{*}\right),\\
Cov^{q(\textbf{h}_{G}^{*})}\left(\textbf{y}^{*}\right)&=Cov^{q(\textbf{h}_{G}^{*})}\left(\textbf{f}_{G}^{*}\right)+\sigma_{G}^{-2}I_{n^{*}}. \label{Pre_mean_covar}
\end{aligned}
\end{equation}

The methodology given above can be used to approximate the predictive density, $p(\textbf{h}_{i}^{*}\mid \textbf{h}_{i}, \textbf{h}_{i-1}^{*},\textbf{h}_{i-1})$ by replacing $\textbf{y}$ with $\textbf{h}_{i}$ and $\textbf{h}_{G}$ with $\textbf{h}_{i-1}$ for $i=2,\ldots,G$. The steps to predict $\textbf{y}^{*}$ given the test input $\textbf{x}^{*}$ for the trained deep GP model (see Equation \eqref{FLB}) can be summarised as follows:
\begin{enumerate}
	\item Let,  $\mathbb{X}$ (inputs), $\textbf{h}_{i}\in \mathbb{R}^{n\times q_{i}}$ (trained hidden values), $\textbf{x}^{*}\in \mathbb{R}^{n^{*}\times \kappa}$ (test inputs); and optimized the corresponding variational and model hyper-parameters
	
	\item Evaluate the predictive mean $\mbox{\boldmath{$\mu$}}_{1}^{*}$ and the predictive variance ${\mbox{\boldmath{$\Sigma$}}}_{1}^{*}$
	
	\item For $i=2:G$, compute the predictive mean $E^{q(\textbf{h}_{i-1}^{*})}\left(\textbf{h}_{i}^{*}\right)$, \& predictive covariance $Cov^{q(\textbf{h}_{i-1}^{*})}\left(\textbf{h}_{i}^{*}\right)$, using the expressions in Equation~(\ref{Pre_mean_covar})
\end{enumerate}
It should be noted that the computation of the predictive means and variances of the model output and each hidden layer is quite straightforward and tractable, since the required statistics can be easily extracted from the variational GP-LVM models. The derived predictive density of each layer output plays an important role in drawing samples from that layer and from the trained deep GP model which will be discussed in the next section.

\section{Applications to uncertainty quantification}\label{sec:uq}

In uncertainty quantification tasks, one specifies a probability density on the inputs $\textbf{x}$ by $\pi(\textbf{x})$, and tries to quantify the probability measure induced by it on the output field, $\textbf{y}=\mathbf{g}(\textbf{x})$, where $\mathbf{g}(\textbf{x})=\mathbf{f}_{G+1}(\mathbf{f}_{G}(\ldots\mathbf{f}_{1}(\mathbf{x}))) $. In this work, a probabilistic generative modelling using the deep GP model is presented. This is an efficient approach in propagating uncertainties from a random and yet uncertain input through nonlinear latent mappings to the outputs as detailed in Section~\ref{sec:DeepGP}. The probability measure obtained in a fully Bayesian setting can be used to estimate the predictive statistics. Moreover, the probability distribution evaluates the lack of information regarding the model output due to the finite number of observations or randomness of the model inputs. Oakley and O'Hagan(2002)\nocite{Oakley.O'Hagan.2002} were among the first researchers who used the surrogate Gaussian process for uncertainty analysis. They proposed a solution to uncertainty analysis of $y=f(\mathbf{x})$ approximated by a GP surrogate model \cite{Oakley.O'Hagan.2002}.
\par
Here, we extend their method to develop an uncertainty analysis method for the deep GP model as introduced earlier.
The proposed procedure is conceptually simple and can be easily developed based on the trained deep GP model. Regarding this, the mean and variance of a function of $\textbf{y}$ can be estimated by sampling from the variational posterior distribution given in Section~\ref{sec:DeepGP}. This approach can be extended to make inferences about the distribution functions of the output $\textbf{y}$. Although, the analytical solution is not always available, the method is quite tight fitted based on the learned variational distributions which can be used to compute the statistics of interests. The proposed approach can be implemented using the following steps:

\begin{enumerate}
	\item Let, the observed $\mathbb{X}$ and $\mathbb{Y}$, and trained deep GP model as explained in Section~\ref{sec:DeepGP}
	
	\item Sample $\mathbb{X}'=\{\mathbf{x}'_{i}\}_{i=1}^{n'}$ from the input distribution $\pi(\mathbf{x})$
	
	\item Generate random data $\mathbf{d}_{1}^{(j)}=\{\mathbf{h}_{11}^{(j)}=\mathbf{f}_{1}^{(j)}(\mathbf{x}_{1}^{'}),\ldots,\mathbf{h}_{1n'}^{(j)}=\mathbf{f}_{1}^{(j)}(\mathbf{x}_{n'}^{'})\}$
	
	\item Approximate $\textbf{f}_{1}^{(j)}(.)$ by the posterior predictive mean ( see Equation~\ref{pre_mean_GP}) of the GP model fitted to $\textbf{d}_{1}^{(j)}$
	
	\item Generate random data $\textbf{d}_{2}^{(j)}=\{\textbf{h}_{21}^{(j)}=\textbf{f}_{2}(\textbf{h}_{11}^{(j)}),\ldots,\textbf{h}_{2n'}^{(j)}=\textbf{f}_{2}(\textbf{h}_{1n'}^{(j)})\}$
	
	\item  Approximate $\textbf{f}_{2}^{(j)}(.)$ by the variational posterior predictive mean of the GP model fitted to $\textbf{d}_{2}^{(j)}$
	
	\item Continue this process to generate random data\\ $\textbf{d}_{g}^{(j)}=\{\textbf{h}_{g1}^{(j)}=\textbf{f}_{g}(\textbf{h}_{(g-1)1}^{(j)}),\ldots,\textbf{h}_{gn'}^{(j)}=\textbf{f}_{g}(\textbf{h}_{(g-1)n'}^{(j)})\}$, and approximate $\textbf{f}_{g}^{(j)}(.)$ using the variational posterior predictive mean of the GP model fitted to $\textbf{d}_{g-1}^{(j)}$.
	
	\item Generate the random data
	\[
	\textbf{d}_{G+1}^{(j)}=\{\textbf{f}_{G+1}^{(j)}(\textbf{h}_{G1}^{(j)}),\ldots,\textbf{f}_{G+1}(\textbf{h}_{Gn'}^{(j)})\},
	\]
	where, $\textbf{f}_{G+1}^{(j)}(.)$ denotes the function that govern the mapping between the model outputs and its dimensionality reduced subspace, $\textbf{h}_{G}$.
	
	\item Approximate $\textbf{f}_{G+1}^{(j)}(.)$ using the posterior mean of the GP model fitted to $\textbf{d}_{L+1}^{(j)}$.
\end{enumerate}

The sample, $\{\mathbf{x}'_{1}.\ldots,\mathbf{x}'_{n'}\}$ drawn from the input distribution $\pi(\mathbf{x})$ are distinct from the original $n$ design points/training data.   
The QoI of $\mathbf{y}$, denoted by $\mathfrak{F}_{j}(Y)$ can now be computed either analytically or numerically. In many situations, $\mathfrak{F}_{j}(Y)$ can be evaluated by solving an integral of multiplications of Gaussian distributions which is trivial. However, this integral generally cannot be analytically computed, and the Monte Carlo method as presented below is then used to compute this statistics:
\[
\hat{\mathfrak{F}}_{j}(Y)=\frac{1}{n'}\sum_{i=1}^{n'}\textbf{f}_{G+1}^{(j)}(\textbf{h}_{Li}^{(j)})
\]
To get a new realisation of $\mathfrak{F}_{j}(Y)$, steps 2--9 are repeated. These realisations, $\{\hat{\mathfrak{F}}_{1}(Y),\ldots,\hat{\mathfrak{F}}_{N'}(Y)\}$ will then form a sample which can be used to estimate ant summary of the distribution of $\mathfrak{F}(Y)$.
\par
Given $\mathbb{X}'$, the posterior predictive distribution of $\mathbf{h}_{1}=\mathbf{f}_{1}$ follows a Gaussian distribution as discussed in Section~\ref{sec:GP_Reg} and presented in Eqs.\ref{pre_dis_GP}--\ref{pre_var_GP}. However,  there is no close form for the posterior predictive distribution of $\mathbf{h}_{g},\\~{2\leq g \leq G+1}$, and this distribution should be variationally approximated as discussed in the previous section. As a result, the computation of $\mathfrak{F}_{j}(Y)$ is not analytically feasible.           
For instance, the mean of the model output $\mathbf{y}$ is defined as 
\[
E^{(j)}(\mathbf{Y})=\int{\mathbf{f}_{G+1}^{(j)}(\mathbf{h}_{G})q(\mathbf{h}_{G})d\mathbf{h}_{G}}.
\]
\noindent
This statistics can be then estimated using the Monte Carlo method as follows:
\begin{equation}
\hat{E}^{(j)}(\mathbf{Y})=\frac{1}{n'}\sum_{i=1}^{n'}\textbf{f}_{G+1}^{(j)}(\textbf{h}_{Gi}^{(j)}).\label{mean_y_jelement}
\end{equation}
By repeating the process above, we can obtain a sample $\{\hat{E}^{(j)}(\mathbb{Y})\}_{j=1}^{N'}$ ($N'$ large) from the distribution of $E(\mathbf{Y})$, where each $\hat{E}^{(j)}(\mathbb{Y})$ is obtained by propagating the uncertainty from the input through the hidden layers to the model output in a probabilistic way. The mean of the derived distribution of $E(\mathbb{Y})$ is called mean of the mean statistics which will be computed for the illustrative examples presented in the next sections. Similarly, the variance of $\mathbf{y}$ can be estimated as follows
\begin{equation}
\widehat{Var}^{(j)}(\mathbb{Y})=\frac{1}{n'}\sum_{i=1}^{n'}(\textbf{f}_{G+1}^{(j)}(\textbf{h}_{Gi}^{(j)})-\hat{E}^{(j)}(\mathbb{Y}))^2,\label{var_y_jelement}
\end{equation}
where, $\hat{E}^{(j)}(\mathbb{Y})$ is given in Equation (\ref{mean_y_jelement}).
Moreover, by repeating the process to compute $\widehat{Var}^{(j)}(\mathbb{Y})$, new  realisations, $\{\widehat{Var}^{(j)}(\mathbb{Y})\}_{j=1}^{N'}$ ($N'$ large) can be generated which are considered as a sample drawn from the distribution of $Var(\mathbb{Y})$. The mean of this distribution (known as mean of the variance) and the corresponding error bars will be computed in the illustrative examples presented in the next sections.

\subsection{Flow through porous media}
\label{sec:pflow}
Following Bilionis et al.~\cite{journals/jcphy/BilionisZKL13}, we consider a single-phase, steady-state
flow through a random permeability field. The spatial domain $\mathcal{X}_s$ is chosen to
be the unit square $[0,1]^2$, representing an idealized oil reservoir. Let us denote
with $p$ and $\textbf{u}=(u_{x},~u_{y})$ the pressure and the velocity fields of the
fluid, respectively. These are connected via the Darcy law:
\begin{equation}
\label{eqn:pflow_darcy}
\textbf{u} = -\textbf{K} \nabla p,; \textrm{in}\;\mathcal{X}_s,
\end{equation}
where $\textbf{K}$ is the permeability tensor that models the easiness with which the liquid flows through the reservoir. Combining the Darcy law with the continuity equation, it is easy to show that the governing PDE for the pressure is:
\begin{equation}
\label{eqn:pflow_pde}
-\nabla\cdot\left(\textbf{K}\nabla p\right) = f,~~ \textrm{in}\;\mathcal{X}_s,
\end{equation}
where the source term $f$ may be used to model injection/production wells.
\par
We use two model square wells: an injection well on the left-bottom corner of $\mathcal{X}_s$ and a production well on the top-right corner.
The particular mathematical form is as follows:
\begin{equation}
\label{eqn:pflow_f}
f(\textbf{x}_{s}) = \begin{cases}
-r,&\;\text{if}\;\;|x_{si}-\frac{1}{2}w| < \frac{1}{2}w,\;\text{for}\;i=1,2,\\
r,&\;\text{if}\;\;|x_{si}-1+\frac{1}{2}w| < \frac{1}{2}w,\;\text{for}\;i=1,2,\\
0,&\;\text{otherwise},
\end{cases}
\end{equation}
where $r$ specifies the rate of the wells and $w$ their size (chosen to be $r=10$ and $w=1/8$). Furthermore, we impose no-flux boundary conditions on the walls of the reservoir:
\begin{equation}
\label{eqn:pflow_no_slip}
\textbf{u} \cdot\hat{\textbf{n}} = 0,\;\text{on}\;\partial\mathcal{X}_s,
\end{equation}
where $\hat{\textbf{n}}$ is the unit normal vector of the boundary. To assure uniqueness of
the boundary value problem defined above we impose the constraint
\[
\int_{\mathcal{X}_s} p(x)dx = 0.
\]
We restrict ourselves to an isotropic permeability tensor:
\begin{equation*}
K_{ij} = K\delta_{ij}.
\end{equation*}
$K$ is modeled as
\[
K(\textbf{x}_s) = \exp\left\{ G(\textbf{x}_s) \right\},
\]
where $G$ is a Gaussian random field:
\[
G(\cdot) \sim \mathcal{N}\left(m, c_{G}(\cdot,\cdot)\right),
\]
with constant mean $m$ and an exponential covariance function given by
\begin{equation}
\label{eqn:pflow_exp_cov}
c_G(\textbf{x}_{s1}, \textbf{x}_{s2}) = s_G^2\exp\left\{ -\sum_{k=1}^{k_s}\frac{|x_{s1,k}-x_{s2,k}|}{\lambda_k} \right\}.
\end{equation}
The parameters $\lambda_k$ represent the correlation lengths of the field, while $s_G>0$ is its variance. The values we choose for the parameters are $m=0$, $\lambda_k=0.01$ and $s_G=1$. In order to obtain a finite dimensional representation of $G$, we employ the Karhunen-Lo\`eve expansion~\cite{Ghanem:1991} and truncate it after $k_{\xi}=50$ terms:
\[
G(\textbf{w}, \textbf{x}_s) = m + \sum_{k=1}^{k_{\xi}}w_k\psi_k(\textbf{x}_s),
\]
where $\textbf{w} = (w_1,\dots,w_{k_{\xi}})$ is vector of independent, zero mean and unit variance Gaussian random variables and $\psi_k(\textbf{x}_s)$ are the eigenfunctions of the exponential covariance given in Eq.~\eqref{eqn:pflow_exp_cov} (suitably normalized, of course).
In order to guarantee the analytical calculation of statistics of the first-order $p$ and $\textbf{u}$, we choose to work with the uniform random variables
\[
\xi_k = \Phi(w_k) \sim \mathcal{U}[0,1],\;k=1,\dots,\textbf{k}_{\xi},
\]
where $\Phi(\cdot)$ is the cumulative distribution function of the standard normal distribution.
Putting it all together, the finite-dimensional stochastic representation of the permeability field is:
\begin{equation}
\label{pflow_K}
K(\mbox{\boldmath{$\xi$}}; \textbf{x}_s) = \exp\left\{m + \sum_{k=1}^{k_{\xi}}\Phi^{-1}(\xi_k)\psi_k(\textbf{x}_s)\right\}.
\end{equation}

In order to make the notational connection with the rest of the paper obvious, let us define the response of the physical model as
\[
\textbf{f} : \mathcal{X} \rightarrow \mathcal{R}^{d},
\]
where $\textbf{x}=(\textbf{x}_{\xi}, \textbf{x}_{s})$ denote the input vector consists of both stochastic input and spatial domain defined over  $\mathcal{X}=\mathcal{X}_{\xi}\times\mathcal{X}_s\in \mathcal{R}^{p}$, $\mathcal{X}_{\xi} = [0,1]^{k_{\xi}}$, and $\mathcal{X}_s$ represents the spatial domain defined as a finite element mesh over $[0,1]^2$, $p$ is the input dimension which is $p=64\times 64$ in this example, and $d$ is the output dimension and set at $d=32\times 32$. In this setting, $\textbf{f}$ can be either  
$p(\textbf{x}_{\xi}, \textbf{x}_{s})$ or $\textbf{u}(\textbf{x}_{\xi}, \textbf{x}_{s})$ which are derived as the solutions of the boundary problem defined by Eqs.~(\ref{eqn:pflow_darcy}),~(\ref{eqn:pflow_pde}) and~(\ref{eqn:pflow_no_slip}) at the spatial point $\textbf{x}_s$ for a permeability field given by Eq.~\eqref{pflow_K}.
\par
We now wish to fit a deep GP with two hidden layers to approximate $\textbf{f}(\mbox{\boldmath{$\xi$}}, \textbf{x}_s)$ based on the data generated by solving the boundary problem given in Eqs.~\eqref{eqn:pflow_darcy},~\eqref{eqn:pflow_pde} and~\eqref{eqn:pflow_no_slip} for the defined permeability field as described above. This boundary value problem is solved using the mixed finite element formulation.
\par
The spatial domain is discretized using $64\times 64$ square grid. We randomly generate {$700$ samples of $\mbox{\boldmath{$\xi$}}$ and use these to create the following data-points $\{\mathbf{K}_{r};, r=1,\ldots,700\}$} using 
Eq.~\eqref{pflow_K}. The eigenfunctions of $\mathbf{K}_{r}$ were calculated and ${\mathbf{K}}_{r}(\mbox{\boldmath{$\xi$}}; \textbf{x}_s)$ were approximated using the KL expansion by truncating it after $k_{\xi}=50$, where $\lambda_k=0.1$ and $s_G=1$. We then solve the PDE described above based for these realizations of the permeability field, $\{\mathbf{K}_{r};, r=1,\ldots,700\}$ to derive the corresponding solutions for velocity $(u_{x}, u_{y})$ and pressure, $p$. {We denote the data-points consists of $\textbf{K}_{r}$ and corresponding solutions for the pressure by $\mathbf{d}_{p}=\{(\mathbf{K}_{r}, p_{r}); r=1,\ldots,700\}$.}
\par
\begin{figure}[H]
	\begin{center}
		\scalebox{0.15}{\includegraphics{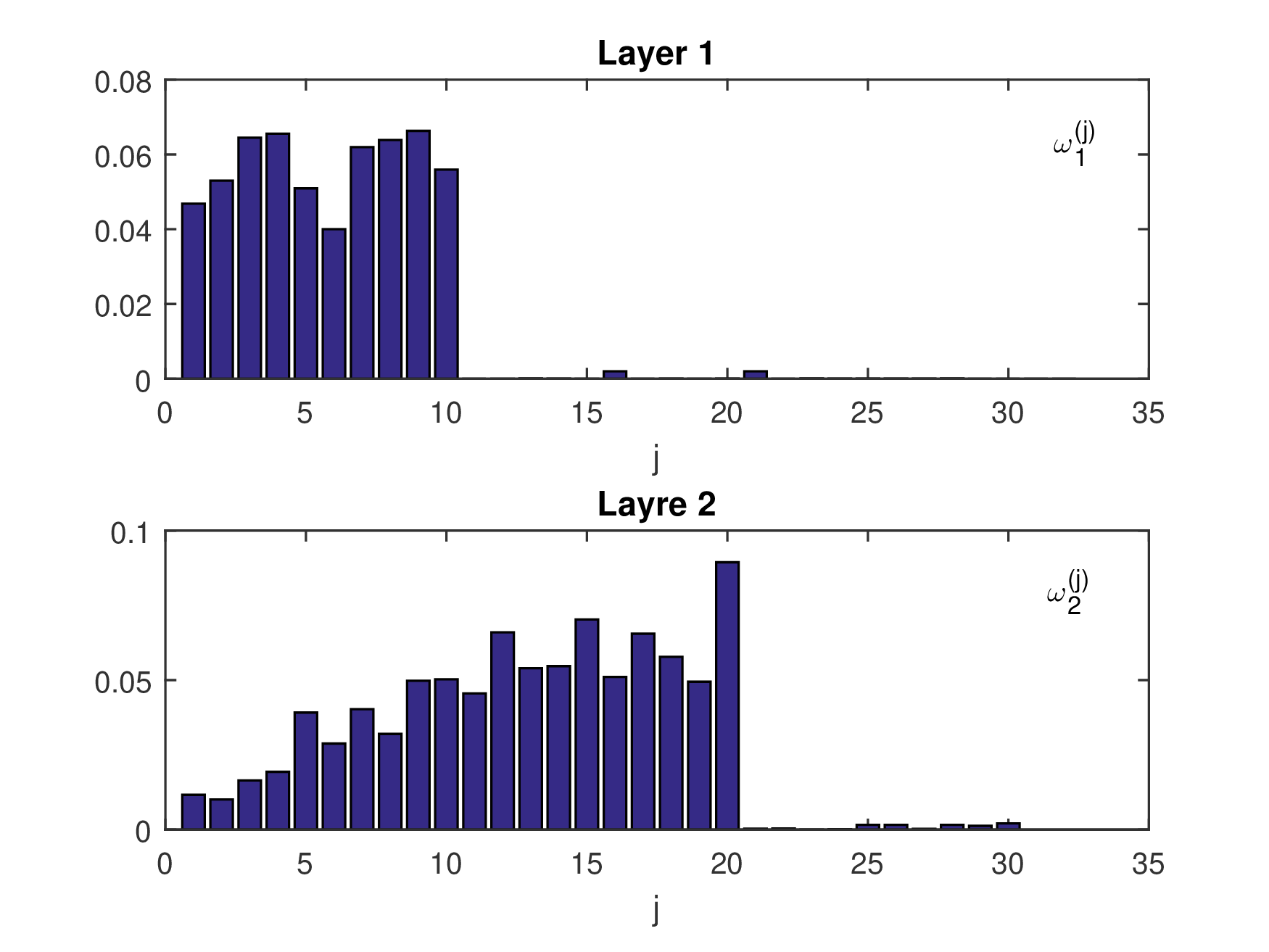}}
	\end{center}
	\caption{The ARD weights of the trained deep GP (with 2 hidden layers fitted to $\mathbf{d}_{p}$) to approximate relationship between the permeability field and the pressure.\label{Flow_Per_p6432}}
\end{figure}

We use the generated data, {$\mathbf{d}_{p}$} for training a deep GP model with two latent variables. {In order to train this model, we use the covariance function given in Eq.~(\ref{Kernel_ARD}) with different length scales. Using this covariance function, an ARD procedure can be implemented to determine the dimensionality of each the latent variables. The length scales ($\omega_{k}$) of this covariance function are automatically estimated during the training the model using the Bayesian variational method as discussed in Section~\ref{sec:Methodology}. The initial guess of the dimensionality of each layer is similarly considered as $q_{1}^{int}=q_{2}^{int}=30$. Based on the details of the trained model, and Fig~\ref{Flow_Per_p6432}, the effective dimensions of the latent layers are determined as, $q_{1}=10$ and $q_{2}=20$.}
\par
\begin{figure}[H]
	\begin{center}
	\begin{subfigure}{6cm}
		\centering\includegraphics[width=5cm]{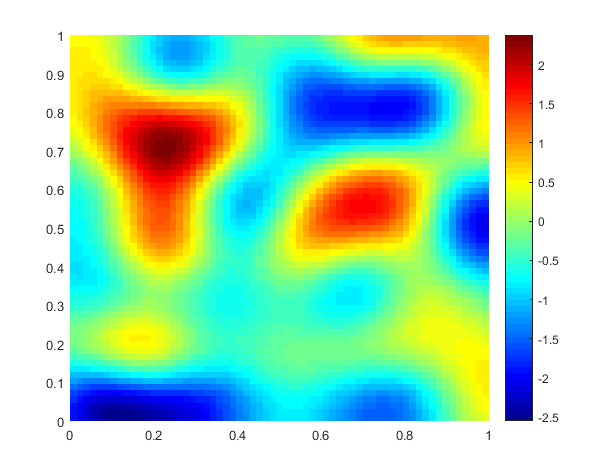}
		\caption{A realisation of Permeability field}
	\end{subfigure}
	\begin{subfigure}{6cm}
		\centering\includegraphics[width=5cm]{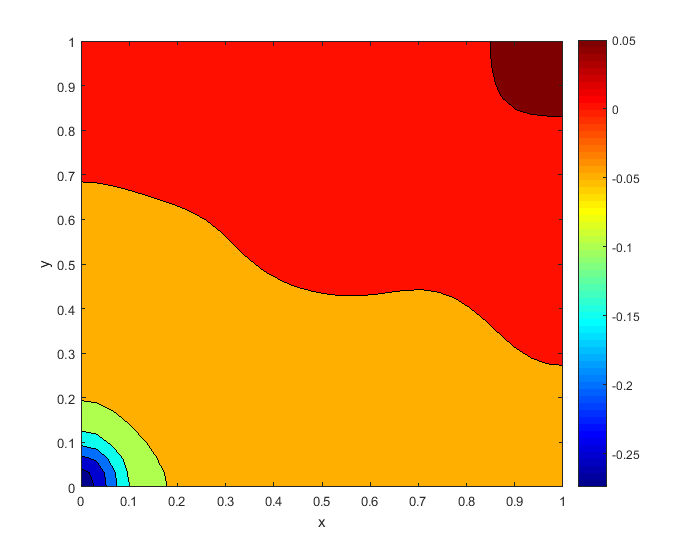}
		\caption{Corresponding solution of $\tilde{\textbf{u}}(\textbf{x})$}
	\end{subfigure}
\caption{(a) The contour plots of the selected test input, $\tilde{\textbf{K}}_{t}(\textbf{x})$ with $\lambda_k=0.1$ and (b) the contour plot of the corresponding solution of $p_{t}$.}\label{Perm_P_6432}
\end{center}
\end{figure}
\begin{figure}[H]
	\begin{center}
	\includegraphics[width=5cm]{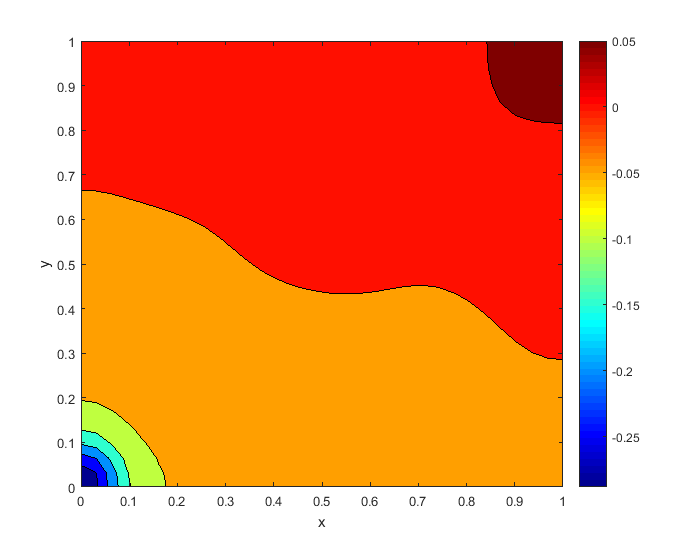}
	\end{center}
	\caption{The contour plot of the posterior predictive mean of $p_{t}$ shown in Sub-figure (b) of Fig.~\ref{Perm_P_6432} .\label{Predict_P_6432_2010}}
\end{figure}

{The contour plots of a test input, $\mathbf{K}_{t}(.)$ and its corresponding solution for the pressure, $p_{t}$ are respectively shown in Sub-figures (a) and (b) of Fig.~\ref{Perm_P_6432}. The corresponding approximation of $p_{t}$ as the posterior predictive mean of the trained deep GP model is shown in Fig~\ref{Predict_P_6432_2010} which is evident that the predictive performance of this surrogate model, despite the limited data used for training the model, is satisfactory.} 
\par

\begin{figure}
	\begin{center}
			\begin{subfigure}{6cm}
				\centering\includegraphics[width=5cm]{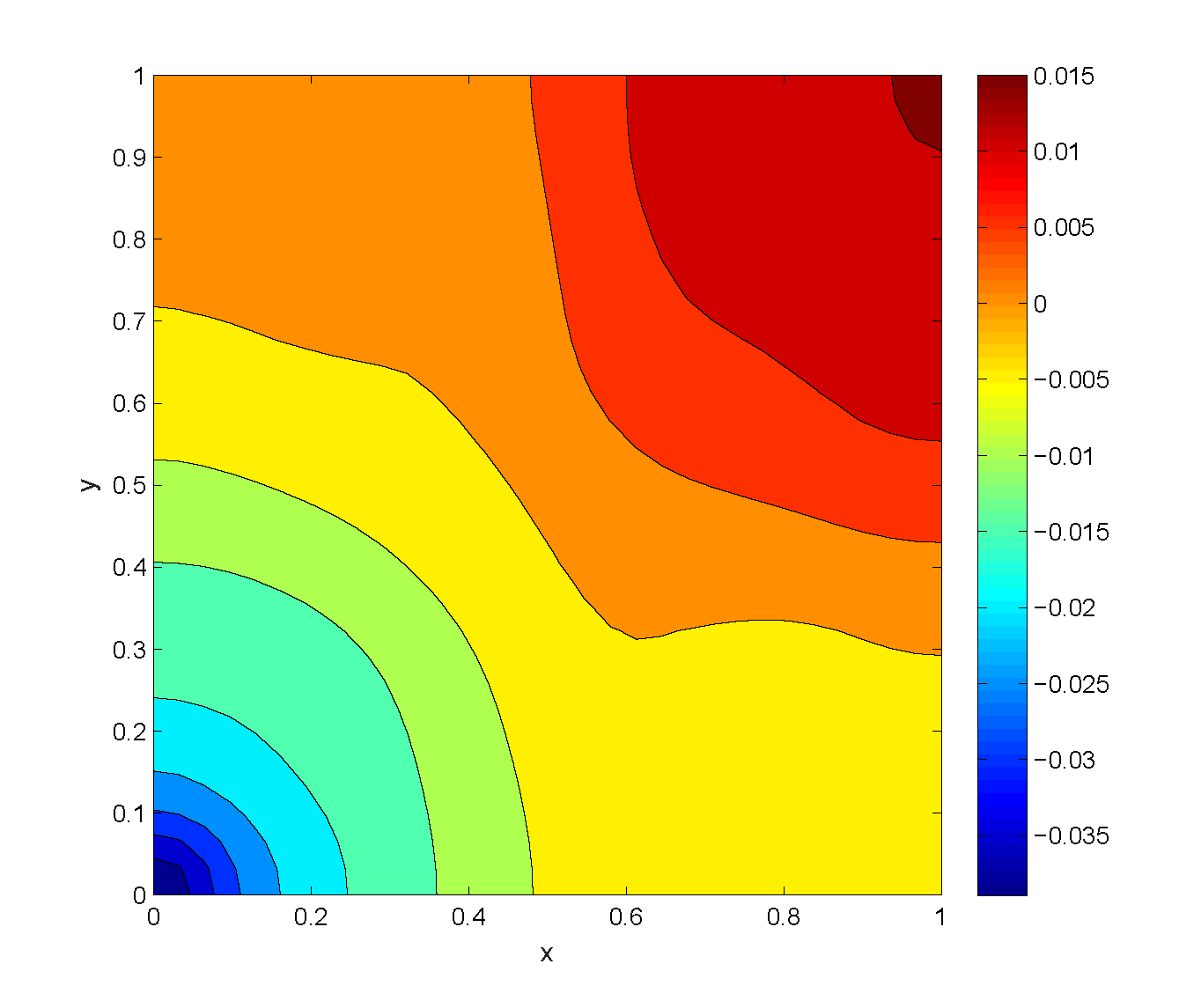}
				\caption{Mean of the mean of $p$, 120 obs.}
			\end{subfigure}
			\begin{subfigure}{6cm}
				\centering\includegraphics[width=5cm]{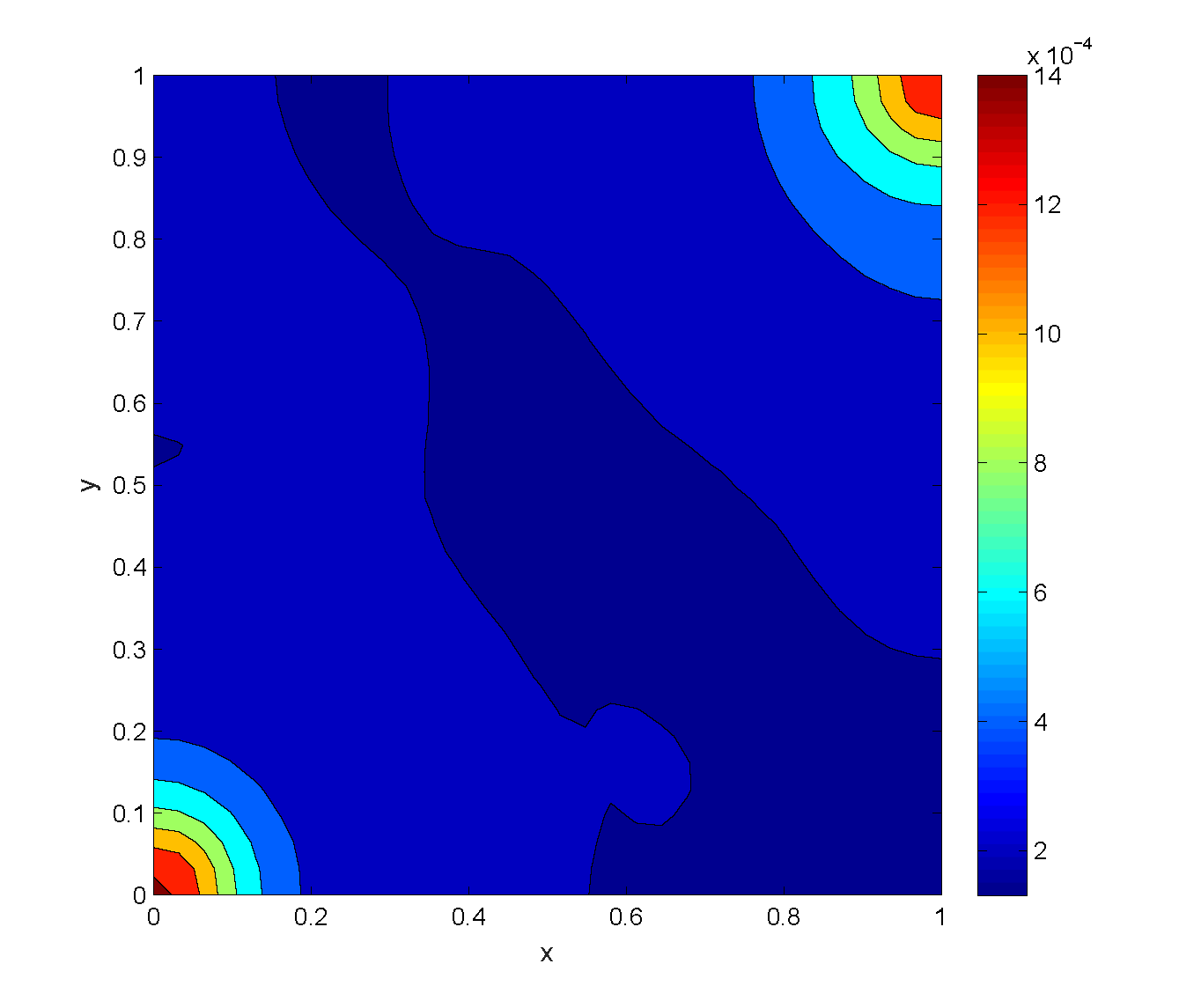}
				\caption{Error bar of the mean of $p$.}
			\end{subfigure}
		\begin{subfigure}{6cm}
			\centering\includegraphics[width=5cm]{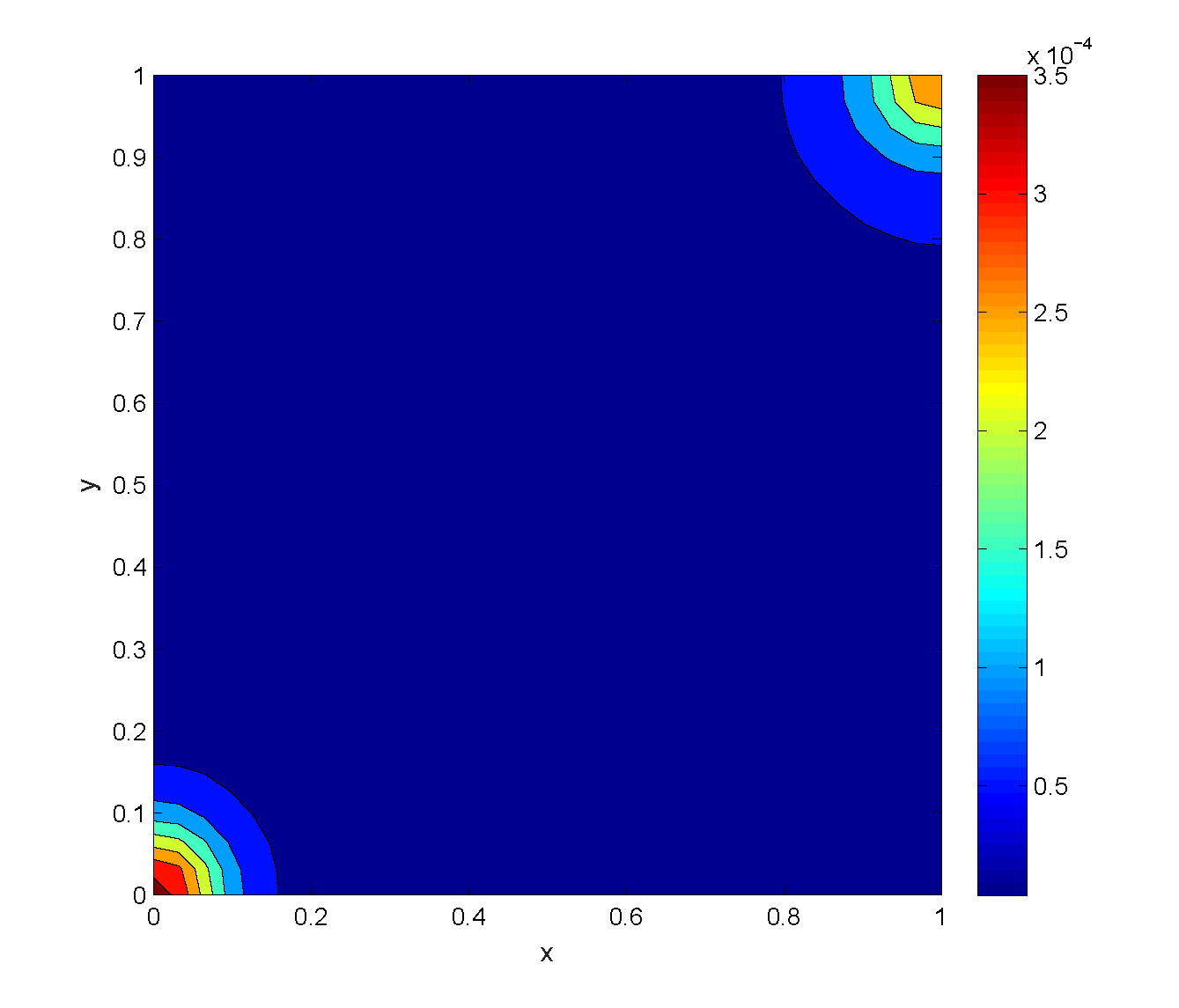}
			\caption{Mean of the variance of $p$, $120$ obs.}
		\end{subfigure}
		\begin{subfigure}{6cm}
			\centering\includegraphics[width=5cm]{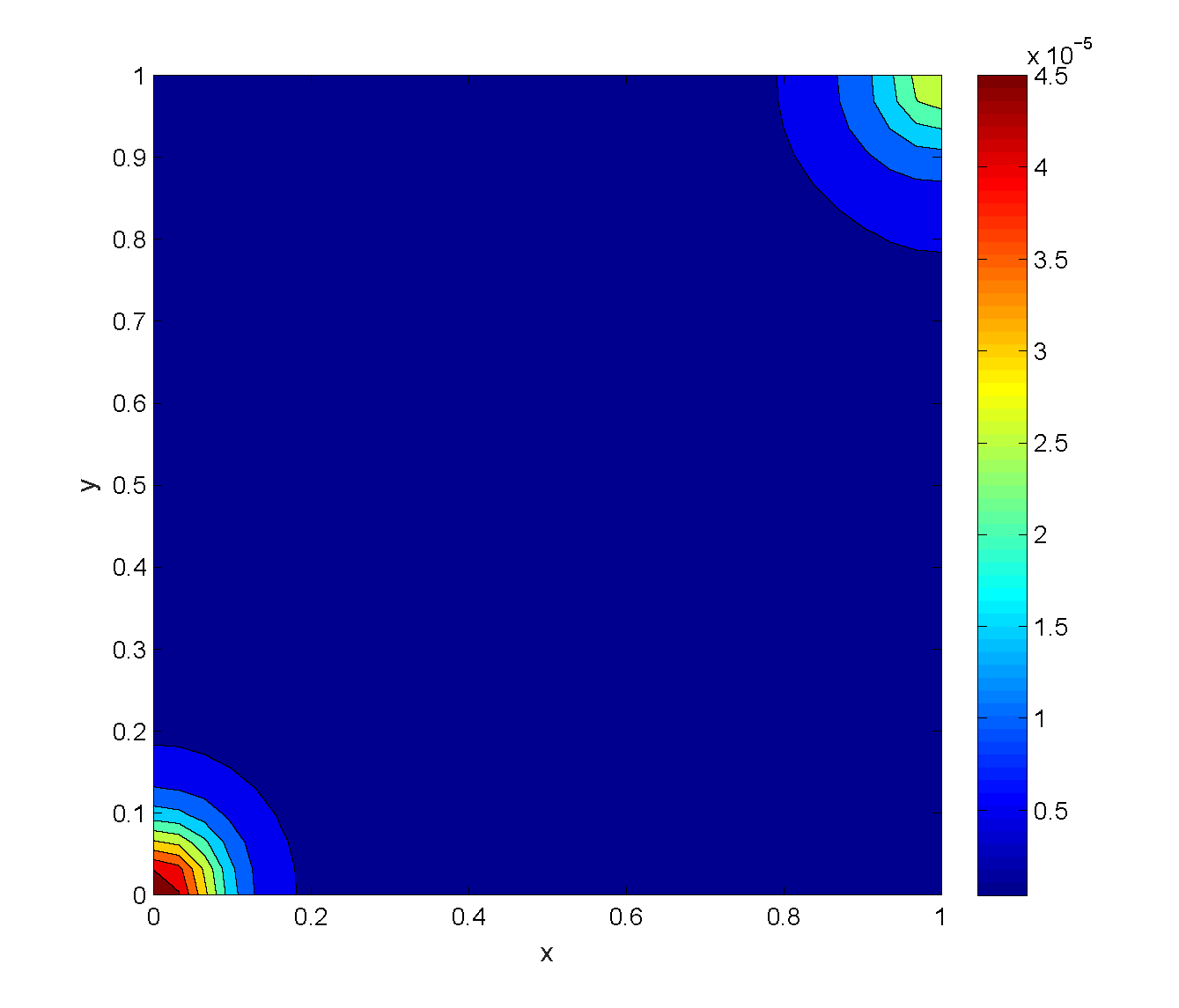}
			\caption{Error bar of mean of variance of $p$.}
		\end{subfigure}
	\caption{Mean of the mean and variance of $p$ evaluated over $32\times 32$ grid using the deep GP model with 120 observations (1st column); and their corresponding error bars (2nd column).}\label{MofM_Error_p}
	\end{center}
\end{figure}

The fitted deep GP model can be used to compute the QoI required to implement UQ task. This model is also very effective for deep learning and probabilistically propagate uncertainty from $\textbf{x}$ to $p$ through two hidden layers. Based on the method discussed Subsection~\ref{sec:uq}, we compute mean of the mean, and mean of the variance of $p$; and their corresponding error bars. Similar to the method described above, we draw a sample of $120$ observations from the input space and compute the corresponding posterior means (and variances) of each latent layer and the model output. By repeating this sampling process for $100$ times, we can compute mean of the mean, mean of the variance and corresponding error bars of the model output. Sub-figure (a) of Fig.~\ref{MofM_Error_p} shows mean of the mean of $p$ using $120$ observations. The corresponding error bar of the mean of $p$ based on $120$ observations is shown in Sub-figures~(b) of Fig.~\ref{MofM_Error_p}. Mean of the variance of $p$ based on $120$  observations is illustrated in Sub-figures (c) of Fig.\ref{MofM_Error_p}. Its corresponding error bar is also shown in Sub-figures (d) of this figure.
\par
\par
\begin{figure}[H]
	\begin{center}
		\scalebox{0.15}{\includegraphics{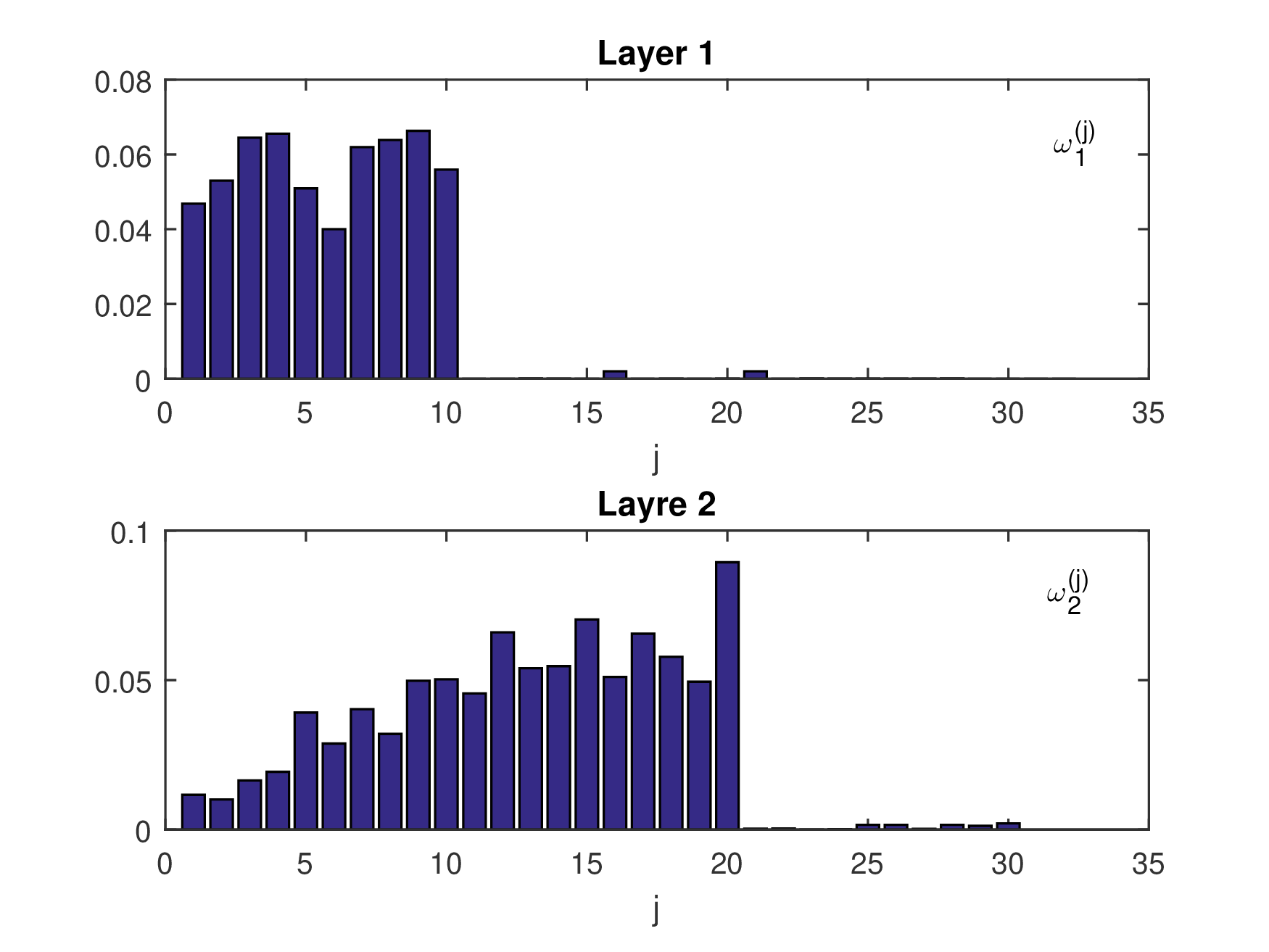}}
	\end{center}
	\caption{The ARD weights of the trained deep GP (with 2 hidden layers fitted to $\mathbf{d}_{x}$) to approximate relationship between the permeability field and $u_{x}$.\label{Flow_Per_ux6432}}
\end{figure}
\begin{figure}[H]
	\begin{center}
		\scalebox{0.65}{\includegraphics{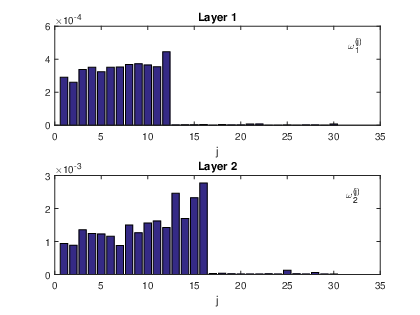}}
	\end{center}
	\caption{The ARD weights of the trained deep GP (with 2 hidden layers fitted to $\mathbf{d}_{y}$) to approximate relationship between the permeability field and $u_{y}$.\label{Flow_Per_uy6432}}
\end{figure}

We now use the same selected realizations of the permeability field,\\ $\{\mathbf{K}_{r};, r=1,\ldots,700\}$ and solve the PDE described in Eqs.~\ref{eqn:pflow_darcy}--\ref{eqn:pflow_no_slip} to derive the solutions for velocity $\mathbf{u}=(u_{x}, u_{y})$. {We denote the data-points consists of $\textbf{K}_{r}$ and corresponding solutions for $u_{x}$ and $u_{y}$, respectively by $\mathbf{d}_{x}=\{(\mathbf{K}_{r}, u_{x}^{r}); r=1,\ldots,700\}$ and  $\mathbf{d}_{y}=\{(\mathbf{K}_{r}, u_{y}^{r}); r=1,\ldots,700\}$. We then fit a two-layer deep GP model to $\mathbf{d}_{x}$ and $\mathbf{d}_{y}$ with the  covariance function given in Eq.~(\ref{Kernel_ARD}) with the different length scales. Figs~\ref{Flow_Per_ux6432} \& \ref{Flow_Per_uy6432} illustrate the ARD weights approximated by fitting the deep GP to $\mathbf{d}_{x}$ and $\mathbf{d}_{y}$, respectively. The effective dimensions of each layer for these two models based on the instruction given above are determined as, $q_{1}=12$ and $q_{2}=16$.}

\begin{figure}[H]
	\begin{center}
		\begin{subfigure}{6cm}
			\centering\includegraphics[width=5cm]{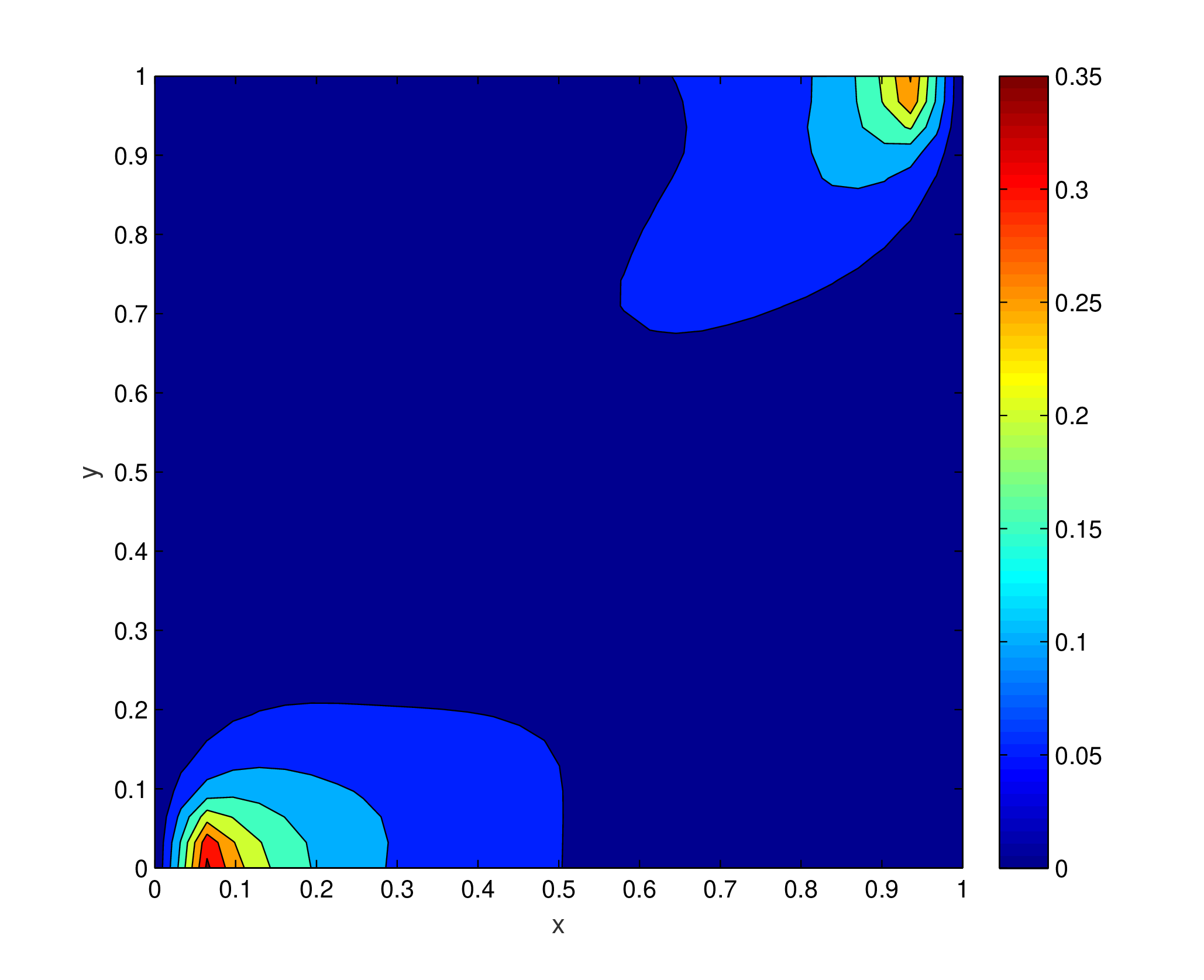}
			\caption{Mean of the mean of $u_x$, $120$ obs.}
		\end{subfigure}
		\begin{subfigure}{6cm}
			\centering\includegraphics[width=5cm]{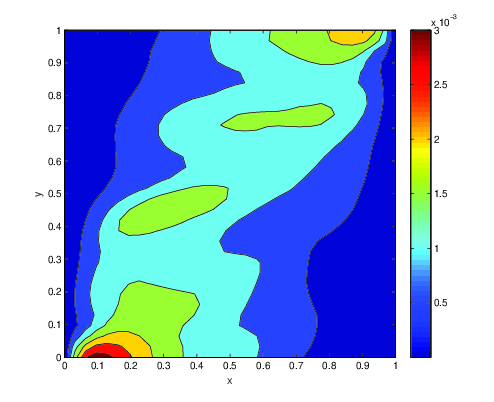}
			\caption{Error bar of mean of $u_x$.}
		\end{subfigure}
		\begin{subfigure}{6cm}
			\centering\includegraphics[width=5cm]{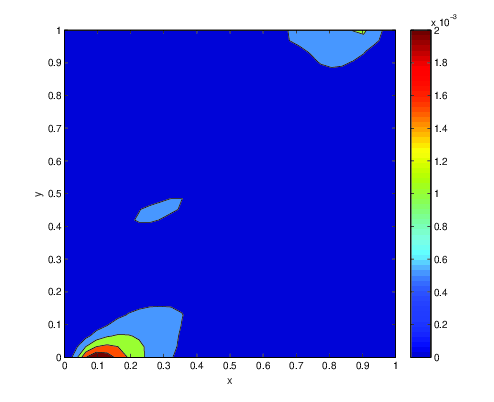}
			\caption{Mean of the variance of $u_x$, $120$ obs.}
		\end{subfigure}
		\begin{subfigure}{6cm}
			\centering\includegraphics[width=5cm]{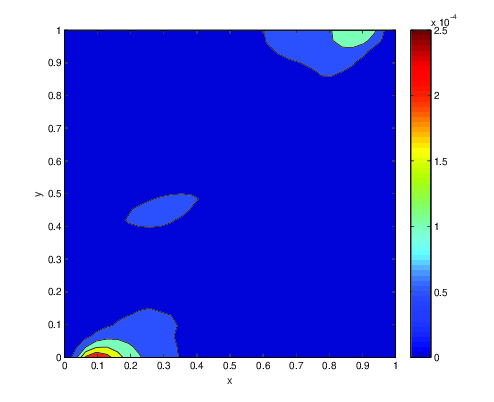}
			\caption{Error bar of variance of $u_x$.}
		\end{subfigure}
	\caption{Mean of mean and variance of $u_x$ evaluated over $32\times 32$ grid using the deep GP with $120$ observations ($1$st column); and corresponding error bars ($2$nd column).}\label{MofM_Error_ux_6432}
	\end{center}
\end{figure}

\begin{figure}[H]
	\begin{center}
		\begin{subfigure}{6cm}
			\centering\includegraphics[width=5cm]{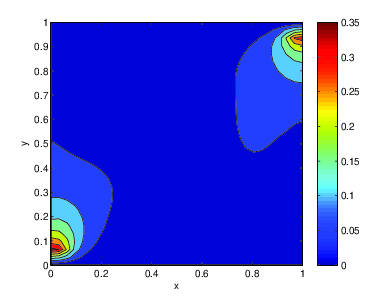}
			\caption{Mean of the mean of $u_y$, $120$ obs.}
		\end{subfigure}
		\begin{subfigure}{6cm}
			\centering\includegraphics[width=5cm]{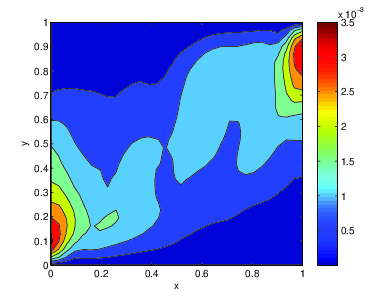}
			\caption{Error bar of mean of $u_y$.}
		\end{subfigure}
		\begin{subfigure}{6cm}
			\centering\includegraphics[width=5cm]{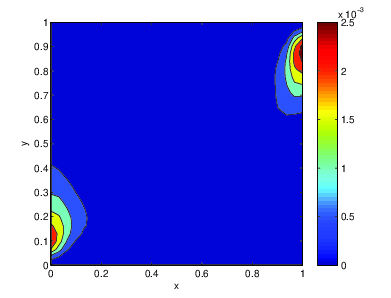}
			\caption{Mean of the variance of $u_y$, $120$ obs.}
		\end{subfigure}
		\begin{subfigure}{6cm}
			\centering\includegraphics[width=5cm]{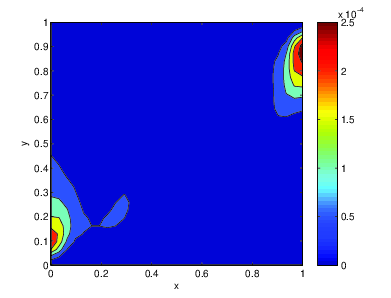}
			\caption{Error bar of variance of $u_y$.}
		\end{subfigure}
\caption{Mean of mean and variance of $u_y$ evaluated over $32\times 32$ grid using the deep GP with $120$ observations ($1$st column); and corresponding error bars ($2$nd column).}\label{MofM_Error_uy_6432}
	\end{center}
\end{figure}

The fitted deep GP models can be then used to compute the QoI required to implement UQ task. We only present the statistics of interest computed from the fitted deep GP model based on the method discussed in Section~\ref{sec:uq}. Fig~\ref{MofM_Error_ux_6432} show means of the mean and means of the variance of $u_{x}$ and their corresponding error bars based on a sample of size $120$. Fig.~\ref{MofM_Error_uy_6432} also show means of the mean and means of the variance of $u_{y}$ and their corresponding error bars based on a sample of size 120. 

\begin{figure}[H]
	\begin{center}
		\begin{subfigure}{6cm}
			\centering\includegraphics[width=5cm]{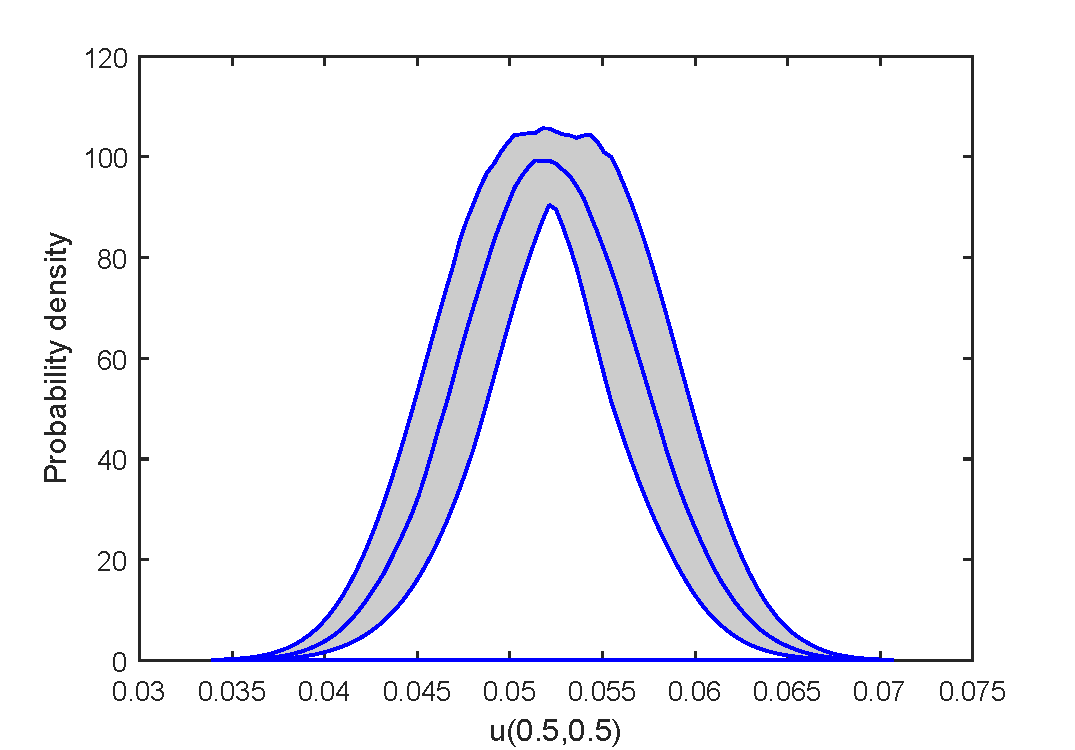}
			\caption{PDF of $\textbf{u}_{x}(0.5, 0.5)$ based on 120 obs.}
		\end{subfigure}
		\begin{subfigure}{6cm}
			\centering\includegraphics[width=5cm]{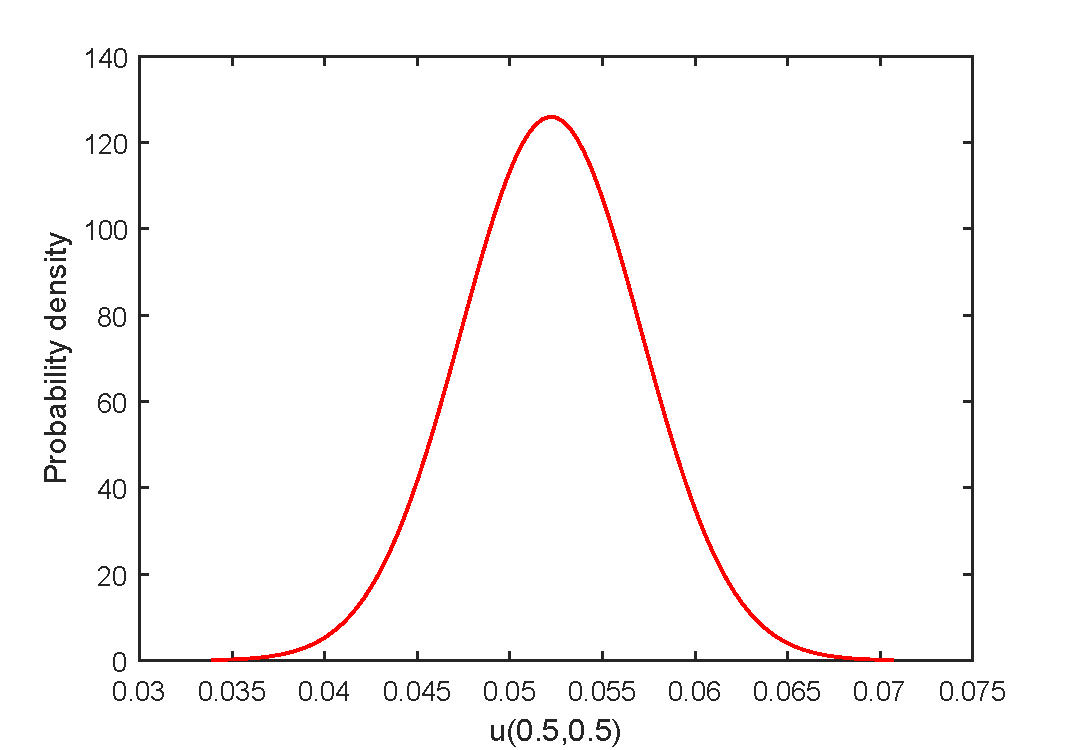}
			\caption{PDF of $\textbf{u}_{x}(0.5, 0.5)$ using MC  Method, 10000 samples.}
		\end{subfigure}
	\caption{The PDF of $\textbf{u}_{x}(0.5, 0.5)$. The middle blue line shows the average PDF over $100$ sampled of the deep GP model for the case of $120$ (a) observations. The filled grey area corresponds to two standard deviations of the PDFs about the mean PDF. The solid red line of (b) is the Monte Carlo estimate using $10000$ observations.}\label{pdf_ux_1612}
	\end{center}
\end{figure}

Fig.~\ref{pdf_ux_1612} (a) sketches the predicted probability density function of $u_{x}(0.05, 0.05)$ along with its error bars, based on 120 data points. The probability density function (PDF) of $u_{x}(0.05, 0.05)$ using the Markov chain method and based on 10000 samples is also illustrated in Fig.~\ref{pdf_ux_1612} (b). In order to compute the PDF shown in Fig.~\ref{pdf_ux_1612} (a), the same training data-points used to compute mean of the mean/variance can be used. However, the MC-based estimated PDF shown in  Fig.~\ref{pdf_ux_1612} (b) required at least 10000 sample sizes. 

\section{Conclusions}\label{sec:con}
We present a novel methodology for efficiently implementing UQ tasks of  complex flow  model which consists of high-dimensional inputs and outputs using the deep GP model. These models generalize several interesting GP-based models, are computationally more efficient, mathematically more elegant, and are more straightforward for the development of uncertainty propagation techniques. We show that several common challenges in UQ can be tackled using this multivariate GP model, including the curse of dimensional and computational complexity in the presence of limited data which is always the case for sunsurface problems. Both of these challenges can be efficiently resolved by breaking down
the emulation into several latent variables which ease the computationally complexity of
the multivariate GP model for emulating high-dimensional inputs and outputs. Each latent variable or reduced dimension sub-space can be then learned using Bayesian GP-LVM which can be considered as a probabilistic non-linear DR approach. The final model can be eventually trained by stacking together the trained sub-models using process composition idea. However, an exact Bayesian inference is not available and tractable for this model, but by using a Bayesian variational approximation framework, the marginal likelihood of model outputs given the inputs can be analytically approximated. In addition, the posterior distribution of each latent variable can be also analytically approximated and used for probabilistically propagating uncertainty through the latent variables. Unlike the Isomap approach and other DR approaches mentioned in Section~\ref{sec:Intro}, the proposed Bayesian framework provides a highly effective probabilistic modelling of high dimensional data which lies on a non-linear manifold (or latent space), and allows to efficiently overcome both the out-of-sample problems and pre-image problems. Another benefit of the proposed variational framework is that the dimensional of each latent variable as well as the model architecture can be automatically determined and tuned without the need for cross-validation.
\par
However, the proposed methodology is very efficient in emulating a model with high-dimensional inputs and outputs and computing the statistics of interests required in UQ tasks, we observe certain aspects that require further investigation. Data scarcity might be resolved automatically during the augmenting each latent mapping with the inducing variables, but it requires further investigation to fully address this issue. The detection and identification of discontinuities in high dimensional space is another challenge that could be efficiently resolved using this methodology. It would be interesting to compare the efficiency of this methodology in tackling this issue with other existing methods including Hierarchical binary 
trees~\cite{journals/tase/MaZ09} generalized polynomial chaos~\cite{journals/jcphy/FooWK08}, adaptive refinement of sparse grids~\cite{journals/jcphy/JakemanAX11,journals/siamsc/GorodetskyM14}.
\par
It would be also interesting to explore how the methodology proposed here could be employed for uncertainty quantification of the flow and transport models which include PDF based mathematical models with multi-scale solutions. Due to the large range of scales in the solutions of these models, it would be extremely challenging to solve them
numerically. Tremendous computational resources are required to solve for the small scales of the solution, which makes it prohibitively expensive. Thus finding an effective equation which governs the large scale solution is very important. In particular, the computational complexity of these model can be increased when the inputs and outputs are high-dimensional~\cite{journals/mmas/KoutsourelakisB11}. Therefore, with the same arguments given in this paper, the current surrogate models including conventional and separable multi-out GP models could not be effective for modelling and computing the statistics required for UQ tasks associated with the multi-scale and high-dimensional PDE-based physical models.



%
%


\appendix
\section{The details of computing $\mbox{\boldmath{$\Phi$}}$ Quantities}
\label{ap:Phi}
In this appendix, we present the details of computing $\mbox{\boldmath{$\Phi$}}$ quantities under the GP prior covariances introduced in Section~\ref{Sec:GP-LVM}. We first represent these quantities as follows
\[
\phi_{0}=\sum_{i=1}^{n} \phi_{0}^{(i)},~\textrm{where}~ \phi_{0}^{(i)}=\int{k(\textbf{h}_{i,:},~\textbf{h}_{i,:})\mathcal{N}(\textbf{h}_{i,:}\mid \mbox{\boldmath{$\mu$}}_{i,:}, \textbf{S}_{i})d\textbf{h}_{i,:}},
\]
\begin{align*}
(\mbox{\boldmath{$\Phi$}}_{1})_{i,k}&=\int{k(\textbf{h}_{i,:},~(\textbf{h}_{u})_{k,:})\mathcal{N}(\textbf{h}_{i,:}\mid \mbox{\boldmath{$\mu$}}_{i,:}, \textbf{S}_{i})d\textbf{h}_{i,:}} ,\\
\mbox{\boldmath{$\Phi$}}_{2}=\sum_{i=1}^{n} \mbox{\boldmath{$\Phi$}}_{2}^{(i)},~\textrm{where}~ (\phi_{2}^{(i)})_{k,k'}&=\int{k(\textbf{h}_{i,:},~(\textbf{h}_{u})_{k,:})k((\textbf{h}_{u})_{k',:},~\textbf{h}_{i,:})\mathcal{N}(\textbf{h}_{i,:}\mid \mbox{\boldmath{$\mu$}}_{i,:}, \textbf{S}_{i})d\textbf{h}_{i,:}}.
\end{align*}
For the ARD exponential quadratic kernel as given in Eq.~\eqref{Kernel_ARD}, the above quantities will become as
\begin{align*}
\phi_{0}&=n\sigma_{h}^{2}, \\
(\mbox{\boldmath{$\Phi$}}_{1})_{i,k}&=\sigma_{h}^{2} \prod_{j=1}^{q}\frac{\exp\left(-\frac{1}{2}\frac{b_{j}(\mu_{i,j}-(h_{u})_{k,j})^{2}}{b_{j}S_{i,j}+1}\right)}{(b_{j}S_{i,j}+1)^{\frac{1}{2}}},\\
\mbox{\boldmath{$\Phi$}}_{2}&=\sigma_{h}^{4} \prod_{j=1}^{q}\frac{\exp\left(-\frac{b_{j}((h_{u})_{k,j}-(h_{u})_{k',j})^{2}}{4}-\frac{b_{j}(\mu_{i,j}-\bar{h}_{:,j})^{2}}{2b_{j}S_{i,j}+1}\right)}{(2b_{j}S_{i,j}+1)^{\frac{1}{2}}},
\end{align*}
where 
\[
\bar{h}_{:,j}=\frac{((h_{u})_{k,j}-(h_{u})_{k',j})}{2}.
\]
These are all the components that are required to compute the variational lower bound given in Eq.~\eqref{LOB_com_fin} for the ARD exponential quadratic kernel. \\
\par

\section{Derivatives with respect to $\mbox{\boldmath{$\theta$}}$ and variational parameters}
\label{ap:DerThetaAndVarPar}
\par
Here we present the expressions for the derivatives of the variational bound Eq.~\eqref{GPLVM_QH} which are required to find the optimized values of the model parameters, $\{\mbox{\boldmath{$\theta$}}_{f},\sigma^{2}_{h}\}$ and variational parameters $\{\mbox{\boldmath{$\mu$}}_{:,j}, \textbf{S}_{j}, \textbf{h}_{u}\}$ using the gradient-based optimization approach.
It would be more convenient to reparameterize $\textbf{S}_{j}$ (for $j=1,\ldots,q$) as follows
\[
\textbf{S}_{j}=(\textbf{K}_{h}^{-1}+\mbox{\boldmath{$\Lambda$}}_{j})
,~~\textrm{and}~~ \mbox{\boldmath{$\mu$}}_{:,j}=\textbf{K}_{h}\bar{\mbox{\boldmath{$\mu$}}}_{:,j},
\]
where $\mbox{\boldmath{$\Lambda$}}_{j}$ is given by
\begin{equation}
\mbox{\boldmath{$\Lambda$}}_{j}=-2\frac{\partial \hat{\mathcal{L}}(q(\textbf{h}))}{\partial \textbf{S}_{j}}~\textrm{and}~ \bar{\mbox{\boldmath{$\mu$}}}_{:,j}=\frac{\partial \hat{\mathcal{L}}(q(\textbf{h}))}{\partial \mbox{\boldmath{$\mu$}}_{:,j}}.
\end{equation}
Here, $\mbox{\boldmath{$\Lambda$}}_{j}=diag(\mbox{\boldmath{$\lambda$}}_{j})$ is a diagonal positive definite matrix of $n$-dimensional vectors, and
\[
\frac{\partial {\mathcal{L}}(q(\textbf{h}))}{\partial\bar{\mbox{\boldmath{$\mu$}}}_{:,j}}=\textbf{K}_{h}\left(\frac{\partial \hat{\mathcal{L}}(q(\textbf{h}))}{\partial \mbox{\boldmath{$\mu$}}_{:,j}}-\bar{\mbox{\boldmath{$\mu$}}}_{:,j}\right).
\]
We first present the derivatives of the lower bound with respect to $\mbox{\boldmath{$\mu$}}_{:,j}$ and $\mbox{\boldmath{$\lambda$}}_{j}$ as follows
\[
\frac{\partial {\mathcal{L}}(q(\textbf{h}))}{\partial\mbox{\boldmath{$\lambda$}}_{j}}=-(\textbf{S}_{j}\circ\textbf{S}_{j})\left(\frac{\partial \hat{\mathcal{L}}(q(\textbf{h}))}{\partial \textbf{S}_{j}}+\frac{1}{2}\mbox{\boldmath{$\lambda$}}_{j} \right),
\]
where `$\circ$' stands for the element-wise product of two matrices.
\par
The partial derivatives given above, $\frac{\partial \hat{\mathcal{L}}(q(\textbf{h}))}{\partial\mbox{\boldmath{$\mu$}}_{:,j}}$ and $\frac{\partial \hat{\mathcal{L}}(q(\textbf{h}))}{\partial \textbf{S}_{j}}$ are given by

\begin{align*}
\frac{\partial \hat{\mathcal{L}}(q(\textbf{h}))}{\partial \mbox{\boldmath{$\mu$}}_{:,j}}&=-\frac{\nu\partial\phi_{0}}{\sigma^{2}_{h}\partial \mbox{\boldmath{$\mu$}}_{:,j}}+
\sigma_{h}^{-2}tr\left(\frac{\partial\mbox{\boldmath{$\Phi$}}_{1}^{T}}{\partial \mbox{\boldmath{$\mu$}}_{:,j}}\mathbb{Y}\mathbb{Y}^{T}\mbox{\boldmath{$\Phi$}}_{1}^{T}\textbf{A}^{-1} \right)\\
&+\frac{1}{2\sigma^{2}_{h}}tr\left(\frac{\partial\mbox{\boldmath{$\Phi$}}_{2}^{T}}{\partial \mbox{\boldmath{$\mu$}}_{:,j}}(\nu\textbf{K}_{uu}^{-1}-\sigma^{2}_{h}\nu\textbf{A}^{-1}-\textbf{A}^{-1}\mbox{\boldmath{$\Phi$}}_{1}^{T}\mathbb{Y}\mathbb{Y}^{T}\mbox{\boldmath{$\Phi$}}_{1}\textbf{A}^{-1})\right),
\end{align*}

\begin{align*}
\frac{\partial \hat{\mathcal{L}}(q(\textbf{h}))}{\partial (\textbf{S}_{j})_{k,l}}&=-\frac{\nu}{2\sigma^{2}_{h}}\frac{\partial\phi_{0}}{\partial (\textbf{S}_{j})_{k,l}}+\sigma_{h}^{-2}tr\left(\frac{\partial\mbox{\boldmath{$\Phi$}}_{1}^{T}}{\partial (\textbf{S}_{j})_{k,l}}\mathbb{Y}\mathbb{Y}^{T}\mbox{\boldmath{$\Phi$}}_{1}\textbf{A}^{-1}\right)\\
&+\frac{1}{2\sigma^{2}_{h}}tr\left(\frac{\partial\mbox{\boldmath{$\Phi$}}_{2}}{\partial (\textbf{S}_{j})_{k,l}}\left(\nu\textbf{K}_{uu}^{-1}-\sigma^{2}_{h}\nu\textbf{A}^{-1}-\textbf{A}^{-1}\mbox{\boldmath{$\Phi$}}_{1}^{T}\mathbb{Y}\mathbb{Y}^{T}\mbox{\boldmath{$\Phi$}}_{1}\textbf{A}^{-1}\right)\right),
\end{align*}
with $\textbf{A}=\sigma_{h}^{-2}\textbf{K}_{uu}+\mbox{\boldmath{$\Phi$}}_{1}$.
\par
The gradient of the lower bound, $\mathcal{B}(q(\textbf{h}))$ with respect to $\mbox{\boldmath{$\theta$}}_{f}$ is given by
\[
\frac{\partial\mathcal{L}(q(\textbf{h}))}{\partial\mbox{\boldmath{$\theta$}}_{f}}=\frac{\partial\hat{\mathcal{B}}(q(\textbf{h}))}{\partial\mbox{\boldmath{$\theta$}}_{f}},
\]
where
\begin{align*}
\frac{\partial \hat{\mathcal{L}}(q(\textbf{h}))}{\partial \mbox{\boldmath{$\theta$}}_{f}}&=-\frac{\nu}{2\sigma_{h}^{2}}\frac{\partial\phi_{0}}
{\partial\mbox{\boldmath{$\theta$}}_{f}}+\sigma_{h}^{-2}tr\left(\frac{\partial\mbox{\boldmath{$\Phi$}}_{1}^{T}}{\partial \mbox{\boldmath{$\theta$}}_{f}}\mathbb{Y}\mathbb{Y}^{T}\mbox{\boldmath{$\Phi$}}_{1}\textbf{A}^{-1}\right)\\
&+\frac{1}{2}tr\left(\frac{\partial \textbf{K}_{uu}}{\partial \mbox{\boldmath{$\theta$}}_{f}}\left(d\textbf{K}_{uu}^{-1}-\sigma_{h}^{2}\nu\textbf{A}^{-1}-\textbf{A}^{-1}\mbox{\boldmath{$\Phi$}}_{1}^{T}\mathbb{Y}\mathbb{Y}^{T}\mbox{\boldmath{$\Phi$}}_{1}\textbf{A}^{-1}-\sigma_{h}^{-2}\nu\textbf{K}_{uu}^{-1}\mbox{\boldmath{$\Phi$}}_{2} \textbf{K}_{uu}^{-1}\right)\right)\\
&+\frac{1}{2\sigma_{h}^{2}}tr\left(\frac{\partial \mbox{\boldmath{$\Phi$}}_{2}}{\partial \mbox{\boldmath{$\theta$}}_{f}}\left(d\textbf{K}_{uu}^{-1}-\sigma_{h}^{2}d\textbf{A}^{-1}-\textbf{A}^{-1}\mbox{\boldmath{$\Phi$}}_{1}^{T}\mathbb{Y}\mathbb{Y}^{T}\mbox{\boldmath{$\Phi$}}_{1}\textbf{A}^{-1}\right)\right).
\end{align*}
The same expression as above can be used for the derivatives with respect to the inducing points, $\textbf{h}_{u}$.
\par
The derivative with respect to $\beta_{h}=\sigma_{h}^{-2}$ is given by
\begin{align*}
\frac{\partial \hat{\mathcal{L}}(q(\textbf{h}))}{\partial \beta_{h}}&=\frac{1}{2}\nu\beta_{h}^{-2} tr(\textbf{K}_{uu}\textbf{A}^{-1})+\frac{1}{2\beta_{h}^{-1}}tr\left(\textbf{K}_{uu}\textbf{A}^{-1}\mbox{\boldmath{$\Phi$}}_{1}^{T}\mathbb{Y}\mathbb{Y}^{T}\mbox{\boldmath{$\Phi$}}_{1}\textbf{A}^{-1}\right)\\
&+\frac{1}{2}\left[\nu\left(tr(\textbf{K}_{uu}^{-1}\mbox{\boldmath{$\Phi$}}_{2})+(n-m)
\beta^{-1}-\phi_{0}\right)+tr\left(\mathbb{Y}\mathbb{Y}^{T}\right)+
tr\left(\textbf{A}^{-1}\mbox{\boldmath{$\Phi$}}_{1}^{T}\mathbb{Y}\mathbb{Y}^{T}
\mbox{\boldmath{$\Phi$}}_{1}\textbf{A}^{-1}\right)\right].
\end{align*}


\end{document}